\PassOptionsToPackage{unicode}{hyperref}
\PassOptionsToPackage{hyphens}{url}
\documentclass[10pt,letterpaper]{article}
\usepackage{amsmath,amssymb}
\usepackage[T1]{fontenc}
\usepackage[utf8]{inputenc}
\usepackage{textcomp}
\usepackage{lmodern}
\IfFileExists{upquote.sty}{\usepackage{upquote}}{}
\IfFileExists{microtype.sty}{\usepackage[]{microtype}\UseMicrotypeSet[protrusion]{basicmath}}{}
\usepackage{xcolor}
\usepackage[margin=1in]{geometry}
\usepackage{longtable,booktabs,array}
\usepackage{graphicx}
\usepackage{calc}
\usepackage{etoolbox}
\makeatletter
\patchcmd\longtable{\par}{\if@noskipsec\mbox{}\fi\par}{}{}
\makeatother
\IfFileExists{footnotehyper.sty}{\usepackage{footnotehyper}}{\usepackage{footnote}}
\makesavenoteenv{longtable}

\usepackage{mathptmx}
\usepackage{titlesec}
\usepackage{fancyhdr}
\usepackage{needspace}
\usepackage{placeins}
\usepackage{enumitem}
\usepackage{ragged2e}
\usepackage{changepage}
\selectfont
\AtBeginEnvironment{longtable}{\small\setlength{\tabcolsep}{2.85pt}}
\titleformat{\section}{\normalfont\bfseries\large}{}{0pt}{}
\titleformat{\subsection}{\normalfont\bfseries\normalsize}{}{0pt}{}
\titleformat{\subsubsection}{\normalfont\bfseries\normalsize}{}{0pt}{}
\titlespacing*{\section}{0pt}{11pt plus 2pt minus 2pt}{5pt}
\titlespacing*{\subsection}{0pt}{10pt plus 2pt minus 2pt}{4pt}
\titlespacing*{\subsubsection}{0pt}{9pt plus 2pt minus 2pt}{3pt}
\setlist[itemize]{leftmargin=1.6em,itemsep=1pt,topsep=3pt,parsep=0pt}
\setlist[enumerate]{leftmargin=1.8em,itemsep=1pt,topsep=3pt,parsep=0pt}
\DeclareUnicodeCharacter{2212}{\ensuremath{-}}
\DeclareUnicodeCharacter{2192}{\ensuremath{\rightarrow}}
\DeclareUnicodeCharacter{2264}{\ensuremath{\leq}}
\DeclareUnicodeCharacter{2265}{\ensuremath{\geq}}
\DeclareUnicodeCharacter{221E}{\ensuremath{\infty}}
\DeclareUnicodeCharacter{2248}{\ensuremath{\approx}}
\DeclareUnicodeCharacter{224D}{\ensuremath{\asymp}}
\DeclareUnicodeCharacter{227A}{\ensuremath{\prec}}
\DeclareUnicodeCharacter{227C}{\ensuremath{\preceq}}
\DeclareUnicodeCharacter{2286}{\ensuremath{\subseteq}}
\DeclareUnicodeCharacter{2208}{\ensuremath{\in}}
\DeclareUnicodeCharacter{2209}{\ensuremath{\notin}}
\DeclareUnicodeCharacter{22A5}{\ensuremath{\perp}}
\DeclareUnicodeCharacter{03B2}{\ensuremath{\beta}}
\DeclareUnicodeCharacter{03B8}{\ensuremath{\theta}}
\DeclareUnicodeCharacter{0394}{\ensuremath{\Delta}}
\DeclareUnicodeCharacter{03B1}{\ensuremath{\alpha}}
\DeclareUnicodeCharacter{03B7}{\ensuremath{\eta}}
\DeclareUnicodeCharacter{03C1}{\ensuremath{\rho}}
\DeclareUnicodeCharacter{03C0}{\ensuremath{\pi}}
\DeclareUnicodeCharacter{03C6}{\ensuremath{\varphi}}
\DeclareUnicodeCharacter{2113}{\ensuremath{\ell}}
\DeclareUnicodeCharacter{25AB}{\ensuremath{\square}}
\DeclareUnicodeCharacter{00A7}{\S{}}
\DeclareUnicodeCharacter{2013}{--}
\DeclareUnicodeCharacter{2014}{---}
\DeclareUnicodeCharacter{2018}{`}
\DeclareUnicodeCharacter{2019}{'}
\DeclareUnicodeCharacter{201C}{``}
\DeclareUnicodeCharacter{201D}{''}
\DeclareUnicodeCharacter{00A0}{~}
\DeclareUnicodeCharacter{00D7}{\ensuremath{\times}}
\DeclareUnicodeCharacter{2026}{\ldots}
\DeclareUnicodeCharacter{2011}{-}
\DeclareUnicodeCharacter{00B7}{\ensuremath{\cdot}}
\IfFileExists{bookmark.sty}{\usepackage{bookmark}}{\usepackage{hyperref}}
\IfFileExists{xurl.sty}{\usepackage{xurl}}{}
\hypersetup{hidelinks,pdfauthor={Jie Wang},pdftitle={Coupled Scaling: A Representational Accessibility Framework for Neural Scaling Laws},pdfsubject={Machine Learning; Neural Scaling Laws; Representation Geometry},pdfcreator={LaTeX}}
\author{}
\date{}
\begin{document}
\begingroup
\begin{center}
{\fontsize{15}{17}\selectfont\bfseries Coupled Scaling: A Representational Accessibility Framework for Neural Scaling Laws\par}
\vspace{8pt}
{\normalsize Jie Wang\renewcommand{\thefootnote}{*}\footnote{LLM assistance disclosure. OpenAI's ChatGPT and Anthropic's Claude assisted with literature discovery, code drafting and debugging, statistical cross-checks, formal presentation, and language revision. The author originated the research question and framework; determined the claims, derivations, and empirical design; independently verified the citations, code, and reported results; and takes full responsibility for the work.}\addtocounter{footnote}{-1}\renewcommand{\thefootnote}{\arabic{footnote}}\par}
{\small School of Civil and Commercial Law, Southwest University of Political Science and Law\par}
{\small September 2026\par}
\vspace{4pt}
{\normalsize\bfseries Abstract\par}
\end{center}
We ask when two learning systems trained on the same task under a common resource protocol should share a scaling rate and when their rates should differ. Coupled Scaling answers this through representational accessibility: the task-relevant geometry that a specified architecture--optimization system can reach and how that geometry is acquired as resources grow. In an orthogonal model, unsupported target energy sets the asymptotic floor, while unacquired supported energy sets the finite-budget residual. When acquisition can skip high-value directions, the largest fully acquired prefix no longer determines a unique exponent. For target powers $a_j\asymp j^{-b}$, $b>1$, and prefix log-growth rate $0<\rho\leq 1$, the sharp attainable interval is $[\rho(b-1),\min\{b-1,\rho b\}]$, and every rate in this interval is realized by a fixed acquisition order on a common support. A rank-window condition identifies when the product law $\alpha=\rho(b-1)$ applies, while a fixed-kernel specialization expresses the same task--system dependence through the task-weighted spectral tail. We then audit released capability trajectories to diagnose measurement and comparison effects and propose a staged test in which independently measured geometry predicts held-out loss, scaling rates, and cross-task reversals.
\endgroup
\vspace{5pt}

\section{1. Introduction}\label{introduction}

Neural scaling laws describe how loss changes with model size, data, and
compute. Their regularities support resource allocation and forecasts of
future capability (Hestness et al., 2017; Kaplan et al., 2020; Hoffmann
et al., 2022). A harder comparative question arises when two learning
systems solve the same task under a common resource protocol: why should
their scaling rates agree, and what would make them diverge? Answering
that question requires a link between the learning curve and the task
structure acquired by each system.

Existing theories supply several pieces of that link. Geometric and
spectral accounts connect learning curves to intrinsic dimension,
eigenspectra, target alignment, and statistical regime (Sharma and
Kaplan, 2022; Maloney et al., 2022; Bahri et al., 2024). Controlled
comparisons show that architecture, feature learning, and
preconditioning can change fitted exponents on the same or closely
matched tasks (Ngo and Ravanbakhsh, 2026; Bordelon et al., 2025; Ramani
and Jain, 2026). Other interventions are well described by common
exponents with system-specific coefficients or resource rescalings
(Bansal et al., 2022; Volkova et al., 2026).

These results leave one comparative object unspecified: the
task-relevant structure that a particular learning system can reach
under a particular resource protocol.

We call that object representational accessibility.\footnote{Cheng et
  al.~(2026) use ``coupling'' for a depth--width path \(L = u(N)\), with
  sample size entering conditions for observable gains. Here the term
  denotes the architecture--optimization system in its relation to
  task-relevant geometry.} For a specified architecture, training
procedure, task, and budget, it records which task-relevant geometry can
be reached and how learning moves through it. The framework separates
three channels: \textbf{support}, which determines what the system can
eventually represent; \textbf{realized geometry}, which describes the
representation produced at a given budget; and \textbf{acquisition},
which records the order and probability with which valuable directions
are learned. Systems with similar unrestricted expressive power can
therefore leave different finite-budget residuals, while systems that
acquire the relevant structure at comparable rates can share a scaling
exponent.

The main mathematical result isolates uneven acquisition. In an
orthogonal model, supported directions are ranked by target power, but
the learner may leave an early gap while acquiring useful directions
beyond it. The first missing rank and the number of acquired
positive-target directions give finite residual bounds. Those bounds
yield a sharp interval of attainable power-law exponents and a
rank-window condition under which completed-prefix growth alone
determines the product-law endpoint. Under a common task-side tail, a
reversal in acquisition-rate advantage across two tasks produces a
reversal in residual-exponent advantage. A fixed-kernel specialization
supplies a complementary route in which eigenvalues determine a soft
acquisition profile. The empirical problem then becomes clear: geometry
must predict something that was not read off the loss curve itself. We
first audit released capability-emergence trajectories to identify
measurement and comparison effects, then propose a staged test that
begins with directly observable ranks in a small system and extends to
held-out geometry-to-scaling prediction in feature-learning networks.

Section 2 develops the comparative motivation. Section 3 presents the
framework and exact results. Section 4 gives the audit and test design,
and Section 5 states the implications and research priorities.

\section{2. From Scaling Mechanisms to Learning-System
Comparisons}\label{from-scaling-mechanisms-to-learning-system-comparisons}

\subsection{2.1 Geometry, Spectra, and Acquisition
Frontiers}\label{geometry-spectra-and-acquisition-frontiers}

Empirical power laws summarize how resources and loss covary;
mechanistic accounts explain why the relation arises in a specified
learning problem.\footnote{Empirical curves can display broken power
  laws, delayed inflections, regime changes, and nonmonotonic
  transitions (Caballero et al., 2023). Exponents in this paper refer to
  a specified resource axis, loss, and regime.} Sharma and Kaplan (2022)
connect parameter scaling to intrinsic dimension under smoothness and
generic-function assumptions. Kernel and random-feature theories derive
learning curves from the spectrum of a data--model operator, the target
coefficients in its eigenbasis, and the sample-size, noise, and
regularization regime (Maloney et al., 2022; Bordelon et al., 2020;
Canatar et al., 2021; Bahri et al., 2024). In each case, the learning
rate already depends on a relation between target and system: the same
target can place different amounts of energy in the easy and hard
directions of different operators.

Acquisition-frontier accounts make the same idea discrete. Zou et
al.~(2026) use a sharp monotone frontier in a ranked Zipfian pattern
space. Song et al.~(2026) infer a moving data-scale cutoff from the tail
of a corpus-intrinsic predictive-contribution spectrum. The basic
calculation is simple. If the first \(k^{\star}(R)\) supported
directions have been acquired at resource level \(R\), the remaining
error is \(\overline{A}\left( k^{\star}(R) \right)\). For
\(a_{j} \asymp j^{- b}\) and \(k^{\star}(R) \asymp R^{\rho}\), the tail
sum gives \(E(R) \asymp R^{- \rho(b - 1)}\).

A frontier tied to target rank describes monotone acquisition.
Comparative settings also admit different supports and interleaved
orders, because one system may skip an early high-value direction while
learning many later ones. The completed prefix records an important
bottleneck but not the whole acquired set. Proposition 2 quantifies how
much rate information the prefix retains and how much improvement can
come from acquisition beyond the first gap, extending the tail--frontier
calculation to learning-system comparisons.

\subsection{2.2 What Changes When the Learning System
Changes?}\label{what-changes-when-the-learning-system-changes}

The evidence points to four distinct channels: architecture can change
structural support or its cost; training can reprioritize directions
within a family; interference can alter how directions coexist; and the
resource path can change the comparison itself.

Architecture studies show why the first channel matters. Tay et
al.~(2023) report architecture-dependent curves and rank changes under
matched language-model pretraining. In neural force fields, task-matched
equivariance changes parameter-, data-, and compute-scaling exponents
(Ngo and Ravanbakhsh, 2026). Hierarchical-language experiments connect
the effect to task structure more directly: locality and weight sharing
align convolutional networks with the generator's hierarchy, producing
faster scaling than Transformers while probes track acquisition of that
hierarchy (Cagnetta et al., 2025). Related work identifies
representation-limited regimes, subspace-recovery transitions, and
operation-specific costs (Defilippis et al., 2026a; Jelassi et al.,
2024; Arora et al., 2024). A resource-matched recipe can still alter
several architectural features at once. Wang et al.~(2026) compare a
sparse MoE Transformer whose middle layers are looped twice with an
unlooped model, closely matching per-token FLOPs, total non-embedding
parameters, and KV-cache size. The looped recipe has a steeper
compute-optimal frontier, with its largest downstream advantage on code
and larger gains for longer samples and more in-context examples.
Because the matched recipes also differ in width, expert count, and
attention configuration, the result identifies the combined
intervention. Geometry measurements are needed to determine which
task-relevant changes accompany the advantage.

Training and optimization provide a second route. Feature learning
improves training-time and compute exponents for hard targets outside
the initial kernel's reproducing space, with little change for aligned
targets (Bordelon, Atanasov, and Pehlevan, 2025). Preconditioning
changes fitted model-size exponents in controlled random-feature
regression (Ramani and Jain, 2026), and trained-weight spectra track
excess-risk regimes in shallow networks (Defilippis et al., 2026b). In
GPT-style models, Jha and Reagen (2026) report optimizer-dependent
effective-rank scaling under a common architecture family, data recipe,
and FFN-width schedule. The AdamW--Dion rank difference remains when
extended training is used to approximately match validation perplexity.
A geometric difference can therefore persist at comparable loss.

The orthogonal model isolates support and acquisition from a third
channel, interference. Liu, Liu, and Gore (2025) derive one scaling
contribution from weak superposition and another from overlap among
representation vectors under strong superposition; open language-model
families display the associated overlap signature. Huang et al.~(2026)
connect rare- and complex-feature retention to capacity allocation,
resource competition, and gradient interference. These results motivate
the broader geometric framework, while the exact calculation below
isolates support and acquisition. The fourth channel is the resource
path. Li et al.~(2026) study domain repetition at fixed tokens per
parameter, so total exposure grows with model size; a separate
fixed-domain-fraction comparison answers a different question. A valid
system comparison specifies how data exposure, compute, optimization
time, and other resources change along the chosen axis. Appendix C.1
records the intervention, controls, reported outcome, and remaining
identification question for the principal studies.

\subsection{2.3 From Mechanism Evidence to a Comparative
Framework}\label{from-mechanism-evidence-to-a-comparative-framework}

The common question is now visible: which parts of a target become
accessible to a particular learning system, and at what resource cost?
Structural support records the directions available to the system.
Realized geometry describes how those directions are represented at a
given budget. Acquisition tracks which task-relevant directions are
learned as the budget grows. Indexing all three by task preserves target
dependence; indexing them by the resource protocol prevents a larger
budget from being confused with a different allocation of that budget.
Coupled Scaling uses representational accessibility to organize these
dependencies and recovers the familiar spectrum-and-alignment
description in a fixed-kernel limit.

The framework also includes a shared-rate null. Liu and Gore (2026)
propose a universality class in which time, width, and depth exponents
are fixed while coefficients vary. Volkova et al.~(2026) obtain more
stable extrapolation by fixing Chinchilla exponents to an AdamW
reference and fitting optimizer-specific resource rescalings; Bansal et
al.~(2022) find stable data-scaling exponents under several
interventions. Exponent changes require a measured acquisition
difference that is large enough and structured appropriately to alter
the rate. Crossing two systems with two tasks makes that dependence
observable: the strongest prediction is a task-conditioned reversal in
which the system with the faster measured acquisition process changes
with the task.

\section{3. The Coupled Scaling
Framework}\label{the-coupled-scaling-framework}

\subsection{3.1 Budget-Relative Representational
Accessibility}\label{budget-relative-representational-accessibility}

At a fixed budget, the first question is practical: which task-relevant
geometries can this architecture and training procedure actually reach
under the stated protocol? Let \(A\) denote an architecture and \(O\)
its training procedure, including parameterization, optimizer, schedule,
and training dynamics. The training problem is
\(P = \left( D_{train}\mathcal{,l} \right)\), with its objective and
preprocessing. A budget envelope \(B\) bounds parameters, data, compute,
and optimization time, and random seeds index stochastic runs within
this specification.

Fix a representation descriptor \(M\), such as a feature, similarity,
covariance, or kernel operator, together with an equivalence relation
\(\sim\) on its outputs. Coordinate changes that preserve the chosen
operator count as equivalent. If \(\tau\) is an admissible run and
\(\theta_{\tau}\) its resulting parameters, define

\[\mathcal{U}_{B}(A,O;P) = \{\tau:\tau\text{ is admissible for }(A,O;P),\ cost(\tau) \preccurlyeq B\},\]

\[\mathcal{R}_{B}^{M, \sim}(A,O;P) = \{\left\lbrack M\left( \theta_{\tau} \right) \right\rbrack_{\sim}:\tau \in \mathcal{U}_{B}(A,O;P)\}.\]

Once \(M\) and \(\sim\) are fixed, \(\mathcal{R}_{B}\) is the set of
geometries reachable within the budget. An implementation specifies the
descriptor, equivalence convention, tolerance, and resource schedule. If
a larger budget preserves all smaller-budget runs, then
\(B_{1} \preccurlyeq B_{2}\) implies
\(\mathcal{R}_{B_{1}} \subseteq \mathcal{R}_{B_{2}}\).

An evaluation task selects the relevant part of this set. We denote the
task-conditioned description by \(P_{T,D}\mathcal{R}_{B}(A,O;P)\), where
\(T\) is the target and \(D\) the evaluation distribution. This notation
keeps the target explicit even when it is encoded in the training
objective. It also separates the three channels used below: support,
realized representation, and acquisition. In a fixed-kernel or suitable
lazy-training limit, these channels are described by available
eigendirections, eigenvalues, and target alignment. Within one
architecture, lazy and rich feature-learning regimes can already realize
different geometries (Chizat et al., 2019; Yang and Hu, 2021).

A reachable set says what can happen. Expected-loss scaling also depends
on how often different outcomes occur. Stochastic comparisons therefore
specify the induced distribution over descriptors or acquisition
profiles, or a justified summary of that distribution. The deterministic
models below isolate support and acquisition; repeated runs restore the
distributional question in empirical work.

With stochastic training, the expected curve is a mixture over realized
geometries. Its apparent rate can differ from that of a typical run and
may be governed by a slowly decaying subset of outcomes. The analysis
therefore separates support, realized geometry, and the distribution
over outcomes.

\subsection{3.2 Finite-Budget Loss and the Asymptotic
Floor}\label{finite-budget-loss-and-the-asymptotic-floor}

Finite-budget loss separates two errors. Some task-relevant structure
may lie outside everything the system can ultimately represent under the
protocol. Other structure may be representable but remain unacquired at
the observed budget. The first source contributes to the asymptotic
floor; the second contributes to finite-budget residual loss.

Let \(B(N)\) specify how the remaining resources change as the chosen
scale coordinate \(N\) grows. In a model-size comparison, for example,
this includes the source of unique examples, cumulative training
presentations, mixture weights, sampling order, and training schedule.
Within-task contrasts share this protocol outside the declared
intervention. Suppressing the task index, define the strict floor by

\[L_{A,O}^{\infty}(D) = \liminf_{N \rightarrow \infty}L(N;A,O,D).\]

When loss converges to this floor with a nonnegative residual, the usual
phenomenological description is

\[L(N;A,O,D) \approx L_{A,O}^{\infty}(D) + C(A,O,D)N^{- \alpha(A,O,D)}.\]

Here \(C\) sets the residual scale and \(\alpha\) describes its decline
along the specified resource axis and fitted regime. Finite-window
curvature or delayed acquisition can obscure the asymptote. For a
nonmonotone curve, compare geometry with loss at the same budgeted
state; a best-so-far value measures a different quantity.

The general floor may contain unsupported target energy, irreducible
noise, or persistent error within the represented subspace. Training can
also change the support eventually reached. The exact model now holds
that last complication fixed: architecture determines support, training
determines acquisition order, and acquired coefficients are initially
exact. This isolates how two systems can have the same asymptotic floor
yet different finite-budget losses and rates.

\subsection{3.3 An Exact Model of Support and
Acquisition}\label{an-exact-model-of-support-and-acquisition}

The exact model has a simple division of labor. Support determines what
can ever be represented. Acquisition order determines what has been
learned by a finite budget. A task-based ranking then lets us compare
acquisition with importance without assuming that the learner follows
that ranking.

Represent the target by orthogonal components. Each squared coefficient
is the target energy lost when that mode is omitted. In this subsection,
\(N\) counts acquisition slots; Appendix B.6 converts slots to
parameters, compute, or time. Let \(f^{\text{*}} \in L^{2}(D)\) and fix
a countable orthonormal basis \(\{\phi_{k}\}_{k \geq 1}\):

\[f^{\text{*}} = \sum_{k}^{}c_{k}\phi_{k},\quad\quad c_{k} = \langle f^{\text{*}},\phi_{k}\rangle,\quad\quad\sum_{k}^{}c_{k}^{2} < \infty.\]

The architecture-family support \(S_{A}\mathbb{\subseteq N}\) is the
union of directions available as modeled capacity grows under the
protocol. A priority order \(\pi_{A,O}\) enumerates that support. For
the asymptotic results, \(S_{A}\) is countably infinite and
\(\pi_{A,O}\mathbb{:N \rightarrow}S_{A}\) is bijective, so every
supported direction is eventually acquired. A nested capacity family
jointly realizes successive prefixes of the order. At budget \(N\),

\[{\widehat{f}}_{N} = \sum_{r = 1}^{N}c_{\pi_{A,O}(r)}\phi_{\pi_{A,O}(r)}.\]

Orthonormality separates the loss into unsupported energy and supported
energy not yet acquired:

\[L\left( {\widehat{f}}_{N} \right) = \underbrace{\sum_{k \notin S_{A}}^{}c_{k}^{2}}_{L_{A}^{\infty}(T)} + \underbrace{\sum_{r > N}^{}c_{\pi_{A,O}(r)}^{2}}_{E_{A,O,T}(N)}.\]

\textbf{Proposition 1 (architecture-dependent strict floor).} Let
\(V_{A} = \overline{span}\{\phi_{k}:k \in S_{A}\}\). Then

\[L_{A}^{\infty}(T) = \parallel \Pi_{V_{A}^{\bot}}f^{\text{*}} \parallel^{2}.\]

The floor is exactly the target energy outside support. Changing the
priority order only rearranges directions inside \(S_{A}\), so it leaves
the floor unchanged. Exhaustive acquisition drives the supported
residual to zero. At a finite budget \(\bar{N}\), however,

\[L\left( {\widehat{f}}_{\bar{N}} \right) = L_{A}^{\infty}(T) + E_{A,O,T}\left( \bar{N} \right),\]

and systems with the same floor can still have different attainable
losses.

To study the rate of that difference, rank supported modes with nonzero
target power independently of acquisition order. Write
\(I_{A,O}(N) = \{\pi_{A,O}(r):1 \leq r \leq N\}\) for the acquired
modes. Enumerate the positive-target supported modes as
\(i_{1},i_{2},\ldots\) in nonincreasing power, resolving ties by a
deterministic rule fixed before the coupling comparison, and define

\[a_{A,T,j} = c_{i_{j}}^{2},\quad\quad K_{A,O,T}(N) = \{ j:i_{j} \in I_{A,O}(N)\},\]

\[J_{A,O,T}(N) = max\{ m \geq 0:\{ 1,\ldots,m\} \subseteq K_{A,O,T}(N)\},\]

\[{\overline{A}}_{A,T}(m) = \sum_{j > m}^{}a_{A,T,j}.\]

The set \(K\) records acquired target ranks. The completed prefix \(J\)
ends immediately before the first missing rank; for example, acquiring
\(\{ 1,2,4,6\}\) gives \(J = 2\). The tail \(\overline{A}(J)\) includes
all energy after that prefix, even energy that has already been
acquired, so it is an upper bound on the residual. The first missing
rank supplies the corresponding lower information.

We use \(u_{N} \asymp v_{N}\) for two-sided bounds by positive constant
factors at sufficiently large \(N\), and \(u_{N} \sim v_{N}\) when their
ratio tends to one.

\textbf{Proposition 2 (acquisition bounds and attainable rates).}
Suppress the fixed coupling and task indices. Let
\(M_{N} = \left| K_{N} \right| \leq N\) count the acquired
positive-target ranks. For an infinite sequence of positive target
powers, every finite acquired set satisfies

\[a_{J_{N}\text{+}1} + \overline{A}\left( M_{N} + 1 \right) \leq E(N) \leq \overline{A}\left( J_{N} \right).\quad\quad(1)\]

Consequently,

\[a_{J_{N}\text{+}1} + \overline{A}(N + 1) \leq E(N) \leq \overline{A}\left( J_{N} \right).\quad\quad(2)\]

For any fixed \(0 \leq J \leq M\), the set
\(K = \{ 1,\ldots,M + 1\} \setminus \{ J + 1\}\) attains the lower
bound in (1). Now assume \(J_{N} \rightarrow \infty\) and that the
following tail and prefix log-rates exist:

\[\gamma = \lim_{m \rightarrow \infty}\frac{-\log \overline{A}(m)}{\log m} \in (0,\infty),\quad\quad\rho = \lim_{N \rightarrow \infty}\frac{\log J_{N}}{\log N} \in (0,1].\]

Then

\[\rho\gamma \leq \liminf_{N \rightarrow \infty}\frac{-\log E(N)}{\log N} \leq \limsup_{N \rightarrow \infty}\frac{-\log E(N)}{\log N} \leq \gamma.\quad\quad(3)\]

If \(\overline{A}\left( J_{N} \right)/E(N) = N^{o(1)}\), the residual
exponent exists and equals \(\alpha = \rho\gamma\). A sufficient
bounded-gain condition is
\(E(N) \geq \eta\overline{A}\left( J_{N} \right)\) for some constant
\(\eta > 0\) and all sufficiently large \(N\). For power-law target
energies \(a_{j} \asymp j^{- b}\), \(b > 1\), the sharper interval is

\[\rho(b - 1) \leq \liminf_{N \rightarrow \infty}\frac{-\log E(N)}{\log N} \leq \limsup_{N \rightarrow \infty}\frac{-\log E(N)}{\log N} \leq \min\{ b - 1,\rho b\}.\quad(4)\]

On a common support containing these positive-target directions and
countably many zero-target directions, every
\(\theta \in \left\lbrack \rho(b - 1),\min\{ b - 1,\rho b\} \right\rbrack\)
is attained by a fixed bijective order with \(J_{N} \asymp N^{\rho}\)
and \(E(N) \asymp N^{- \theta}\). Under bounded gain and
\(J_{N} \asymp N^{\rho}\), the product endpoint has the stronger order
statement \(E(N) \asymp N^{- \rho(b - 1)}\).

Equation (1) has a direct reading. The first missing mode contributes
unavoidable error. Even if all remaining slots are allocated to the best
available ranks, a tail remains beyond those ranks. The other extreme
leaves the full tail beyond the completed prefix. When \(\rho = 1\),
both restrictions give exponent \(b - 1\); when prefix growth is slower,
acquisition beyond the first gap can improve the rate. Appendix B.2
proves sharpness and constructs one fixed order for every attainable
exponent.

\includegraphics[width=6.25in,height=4.41715in]{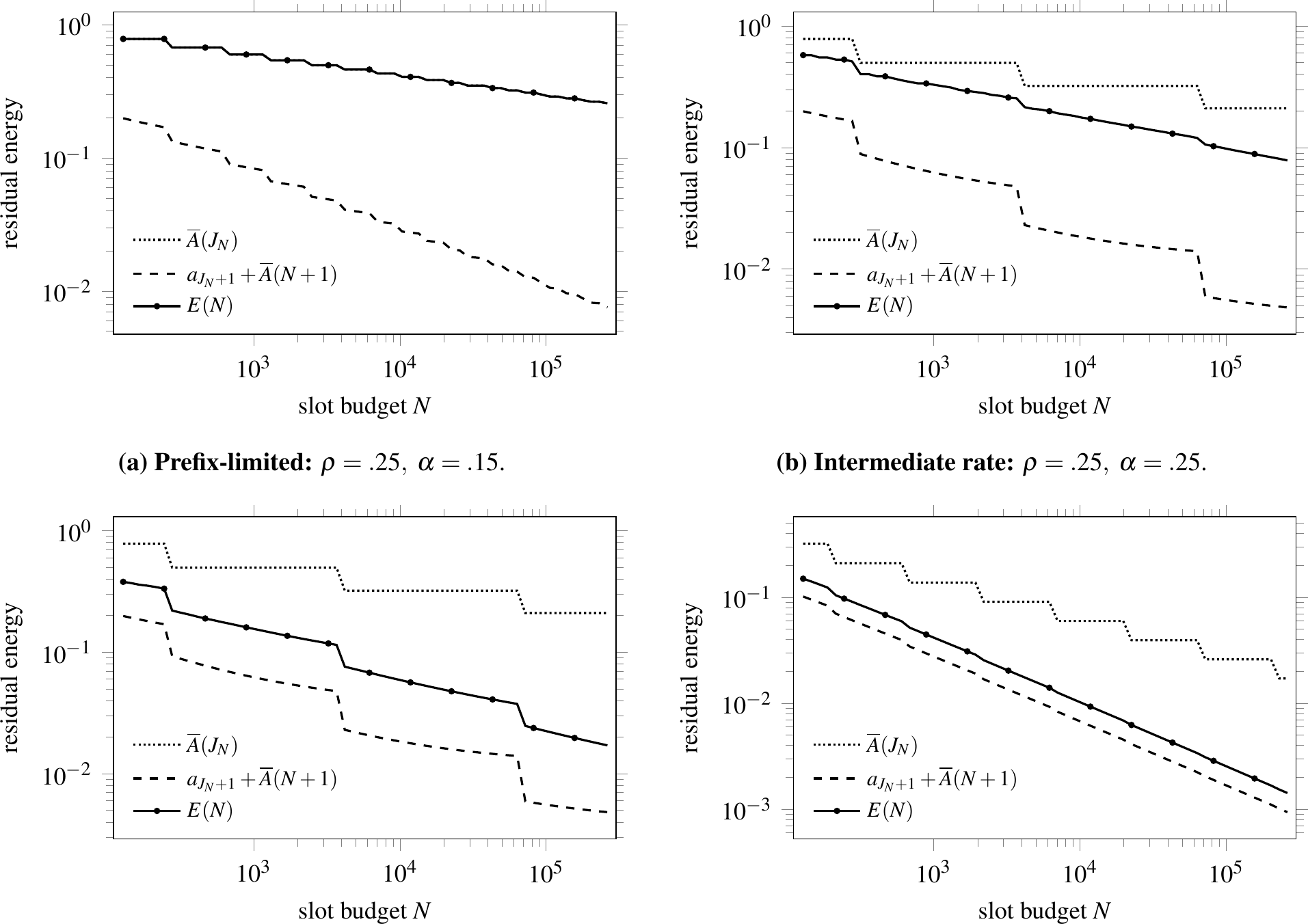}

\textbf{Figure 1. Attainable rates for} \(a_{j} = j^{- 1.6}\)\textbf{.}
Solid marked curves show \(E(N)\), dotted curves
\(\overline{A}\left( J_{N} \right)\), and dashed curves the lower bound
\(a_{J_{N}\text{+}1} + \overline{A}(N + 1)\). With \(\rho = .25\),
panels (a)--(c) attain the lower endpoint .15, an interior rate .25, and
the upper endpoint .40; panel (d) attains rate .60 with \(\rho = .60\).
The fitted upper-half slopes are .149, .254, .397, and .605. The panels
show that the same prefix rate can support different residual rates when
acquisition beyond the first gap differs.

The product endpoint is the most interpretable rate prediction, but it
requires the completed prefix to capture the relevant residual order.
The following condition checks that requirement from unresolved ranks
near the frontier.

\textbf{Lemma 1 (unresolved-rank window).} Suppose
\(a_{j} \asymp j^{- b}\), \(b > 1\), and \(J_{N} \rightarrow \infty\).
If constants \(\lambda > 1\) and \(c > 0\) satisfy

\[\text{\#}\left( \{ J_{N} + 1,\ldots,\lfloor\lambda J_{N}\rfloor\} \setminus K_{N} \right) \geq cJ_{N},\quad\quad(5)\]

for all sufficiently large \(N\), then
\(E(N) \asymp \overline{A}\left( J_{N} \right)\). A proportional number
of unresolved directions in this window retains enough energy to make
the prefix adequate for the exponent. The condition is observable in a
rank-resolved acquisition profile; Appendix B.2 gives the proof.

A cross-system comparison also needs a comparable task tail. Because the
positive-target sequence is ranked after restriction to \(S_{A}\),
\(\gamma_{A,T}\) and \(b_{A,T}\) generally depend on both architecture
and task. One sufficient route is common full support in a task-side
basis fixed independently of the couplings. Under that condition, the
product law transfers a within-task ordering of acquisition rates to the
residual exponents.

\textbf{Corollary 1 (coupling-by-task reversal).} Consider two couplings
\(q_{1} = \left( A_{1},O_{1} \right)\) and
\(q_{2} = \left( A_{2},O_{2} \right)\) and two tasks \(T_{1},T_{2}\).
Within each task, suppose both couplings have full support in the same
ex ante task-side basis, share a cumulative supported-tail log-rate
\(\gamma_{t} > 0\), and satisfy the bounded or subpolynomial off-prefix
condition in Proposition 2. If

\[\rho_{q_{1},T_{1}} > \rho_{q_{2},T_{1}},\quad\quad\rho_{q_{1},T_{2}} < \rho_{q_{2},T_{2}},\]

their residual-exponent ordering reverses across the tasks. The
residual-loss ordering also reverses for all sufficiently large \(N\);
equal within-task floors extend the conclusion to total loss. Common
tails and equal prefix rates give the shared-exponent case within the
same model.

The exact coefficients assumption can be relaxed. Estimation adds a
nonnegative term \(Q(N)\), so
\(L\left( {\widehat{f}}_{N} \right) = L_{supp} + E(N) + Q(N)\). Appendix
B.6 derives this decomposition and the slot-to-resource conversion.
These are the conditions needed to carry an acquisition-rate prediction
to a measured total-loss curve.

\subsection{3.4 Learning Rates in a Fixed-Kernel
Model}\label{learning-rates-in-a-fixed-kernel-model}

The orthogonal model specifies support and acquisition order directly. A
fixed-kernel model supplies a complementary case in which the learning
dynamics determine how quickly each direction is resolved. Target
coefficients place energy in eigenfunctions; eigenvalues set their decay
rates under gradient flow. We use a noiseless population setting to keep
that relation explicit, while standard kernel and random-feature
theories add sample size, noise, and regularization (Bordelon et al.,
2020; Canatar et al., 2021; Caponnetto and De Vito, 2007; Bahri et al.,
2024).

The spectral statement separates any atom at zero as the strict floor.
On the positive spectrum, the task-weighted spectral measure is locally
finite and regularly varying at zero with index β \textgreater{} 0. The
residual is then a Laplace--Stieltjes transform, and Karamata's theorem
(Bingham et al., 1987) gives the stated $t^{-\beta}$ rate up to the slowly
varying factor.

Let \(q\) index a bounded, positive semidefinite, self-adjoint operator
\(K_{q}\) on \(L^{2}(D)\), with the optimizer and population
gradient-flow dynamics fixed. Assume a complete countable orthonormal
eigenbasis \(\{\varphi_{q,k}\}_{k}\), including the zero-eigenvalue
subspace. Write \(\lambda_{q,k} \geq 0\) for its eigenvalues and
\(c_{q,T,k} = \langle f_{T},\varphi_{q,k}\rangle\) for a nonzero target
\(f_{T} \in L^{2}(D)\). Define the normalized task-weighted spectral
measure

\[\nu_{q,T} = \sum_{k}^{}\frac{c_{q,T,k}^{2}}{\sum_{j}^{}c_{q,T,j}^{2}}\,\delta_{\lambda_{q,k}}.\]

This measure assigns target energy to learning rates before a scaling
curve is fitted. Starting from zero, the residual obeys
\(\dot{r} = - K_{q}r\), so \(r(t) = e^{- tK_{q}}r(0)\) and

\[\frac{L_{q,T}(t)}{L_{q,T}(0)} = \int e^{- 2\lambda t}\, d\nu_{q,T}(\lambda),\quad\quad H_{q,T}(t) = 1 - \int e^{- 2\lambda t}\, d\nu_{q,T}(\lambda).\]

Energy at eigenvalue zero persists. Positive eigenvalues decay, and
smaller eigenvalues learn more slowly. The filter \(e^{- 2\lambda t}\)
is therefore a soft acquisition profile, while \(H_{q,T}(t)\) is the
fraction of target energy resolved by time \(t\). Long-time scaling is
governed by the task-weighted mass near zero.

\textbf{Proposition 3 (classical fixed-kernel spectral-tail relation).}
Let \(m_{q,T} = \nu_{q,T}\left( \{ 0\} \right)\) and
\(F_{q,T}(x) = \nu_{q,T}(\left( 0,x\text{]} \right)\). Suppose
\(F_{q,T}\) is regularly varying at zero with index \(\beta_{q,T} > 0\):

\[\lim_{x \downarrow 0}\frac{F_{q,T}(ax)}{F_{q,T}(x)} = a^{\beta_{q,T}}\quad(a > 0).\]

As \(t \rightarrow \infty\), the residual satisfies

\[\frac{L_{q,T}(t) - L_{q,T}^{\infty}}{L_{q,T}(0)} \sim \Gamma\left( \beta_{q,T} + 1 \right)F_{q,T}\left( (2t)^{- 1} \right),\quad\quad L_{q,T}^{\infty} = m_{q,T}L_{q,T}(0).\]

In particular, if
\(F_{q,T}(x) \sim c_{q,T}x^{\beta_{q,T}}\mathcal{l}_{q,T}(1/x)\) for a
slowly varying \(\mathcal{l}_{q,T}\), then

\[L_{q,T}(t) - L_{q,T}^{\infty} \sim L_{q,T}(0)c_{q,T}\Gamma\left( \beta_{q,T} + 1 \right)(2t)^{- \beta_{q,T}}\mathcal{l}_{q,T}(2t).\]

Thus the training-time exponent is \(\alpha_{t}(q,T) = \beta_{q,T}\),
with the stated slowly varying correction. A specified near-zero
spectral tail implies the loss asymptotic; the reverse inference
requires the corresponding Tauberian conditions. A positive spectral gap
produces exponential decay. Appendix B.4 records the short Karamata
derivation.

The product-law connection appears when spectral strength and target
power have aligned rankings. Order the positive eigenvalues as
\(\lambda_{1} \geq \lambda_{2} \geq \cdots \downarrow 0\) and let
\(w_{j} = c_{q,T,j}^{2}/\sum_{k}^{}c_{q,T,k}^{2}\), including
zero-eigenvalue target energy in the denominator. If
\(\lambda_{j} \sim \kappa j^{- a}\) and \(w_{j} \sim dj^{- b}\), with
\(\kappa,d > 0\), \(a > 0\), and \(b > 1\), then

\[F(x) \sim \frac{d}{b - 1}\kappa^{- (b - 1)/a}x^{(b - 1)/a}.\]

The filter resolves modes on the scale \(t\lambda_{j} \gtrsim 1\), so
\(J_{eff}(t) \asymp t^{1/a}\) and \(\beta = (b - 1)/a\). This is the
product of the tail analogue \(\gamma = b - 1\) and the soft-frontier
analogue \(\rho_{eff} = 1/a\). The hard-slot model also imposes
\(J_{N} \leq N\); identifying \(N\) with \(t\) under that normalization
requires \(a \geq 1\). When eigenvalue and target-power rankings are
misaligned, the joint measure \(F\) retains the relevant information.
Opposite \(\beta\) orderings across two tasks again imply eventual
residual-order reversal when within-task floors are equal.

\subsection{3.5 Measuring the Theoretical
Objects}\label{measuring-the-theoretical-objects}

Empirical work can follow two routes. A \textbf{direct route} uses a
controlled system in which target ranks and acquired directions are
observable. A \textbf{proxy route} measures a prespecified aspect of
geometry in a deep network and asks whether it improves held-out
prediction. Table 1 separates the exact theoretical objects from these
empirical roles.

\textbf{Table 1. Theoretical objects and their empirical roles.}

\begin{longtable}[]{@{}
  >{\raggedright\arraybackslash}p{(\columnwidth - 6\tabcolsep) * \real{0.2500}}
  >{\raggedright\arraybackslash}p{(\columnwidth - 6\tabcolsep) * \real{0.2500}}
  >{\raggedright\arraybackslash}p{(\columnwidth - 6\tabcolsep) * \real{0.2500}}
  >{\raggedright\arraybackslash}p{(\columnwidth - 6\tabcolsep) * \real{0.2500}}@{}}
\toprule\noalign{}
\begin{minipage}[b]{\linewidth}\raggedright
Object
\end{minipage} & \begin{minipage}[b]{\linewidth}\raggedright
What it records
\end{minipage} & \begin{minipage}[b]{\linewidth}\raggedright
Candidate measurement
\end{minipage} & \begin{minipage}[b]{\linewidth}\raggedright
Interpretation
\end{minipage} \\
\midrule\noalign{}
\endhead
\bottomrule\noalign{}
\endlastfoot
\(\mathcal{R}_{B}(A,O;P)\) & Geometries feasible within a budget &
Prespecified descriptor projections & Full-set recovery unspecified \\
\(P_{T,D}\mathcal{R}_{B}\) & Task-relevant part of the geometry &
Cross-system, task-conditioned probes & Partly measurable \\
\(g_{q,t}^{level}\left( \bar{N} \right)\) & Breadth or alignment at one
budget & Held-out probes, effective rank, alignment, spectral summaries
& Predicts \(L_{q,t}\left( \bar{N} \right)\) only after proxy
validation \\
\(\gamma_{A,T}\) & Energy remaining beyond target rank & Ex ante task
basis and supported target powers & Exact in the model; basis and
support dependent \\
\(\rho_{A,O,T}\) & Growth of the completed prefix & Rank-resolved
\(J_{q,t}^{(g)}(N)\) & Product prediction requires prefix adequacy \\
\(\beta_{q,T}\) & Task-weighted mass near eigenvalue zero & Near-zero
tail of \(F_{q,T}\) & Exact rate under Proposition 3 \\
Saturation & Possible bottleneck & Plateau, curvature, regime transition
& Diagnostic; needs independent explanation \\
\end{longtable}

In the direct route, a rank-resolved profile yields \(J_{q,t}^{(g)}(N)\)
and the unresolved directions around it. Across budgets, estimate

\[J_{q,t}^{(g)}(N) \asymp N^{\rho_{q,t}^{(g)}}.\]

The target-side powers give the tail rate, while Lemma 1 or an
independent bounded-gain argument determines whether the product
endpoint applies. Substantial acquisition beyond the first gap is
compared with the full interval in Proposition 2. Mode-resolved
experiments can also test the finite bound in (1) directly, avoiding an
unstable asymptotic fit over a short resource window.

In the proxy route, a scalar \(g_{q,t}^{level}\left( \bar{N} \right)\)
summarizes task-relevant breadth or alignment at one budget. Candidate
measurements include adaptation subspaces (Aghajanyan et al., 2021),
activation-space intrinsic dimension (Ansuini et al., 2019), and
effective alignment measures such as \(d_{align}\) (Zhang et al., 2026).
Fix the prediction direction, probe population, layer, normalization,
and target distribution before evaluating loss. Repeated seeds
characterize the resulting descriptor distribution.

A justified mapping to target ranks, acquisition, or a spectral tail
gives a proxy its theoretical interpretation. Absent that mapping, the
empirical question is whether independently measured geometry adds
held-out predictive information beyond a resource-only curve.
Fixed-kernel studies can answer the corresponding question directly
through \(\nu_{q,T}\) and its near-zero tail index \(\beta_{q,T}\);
evolving-representation studies may also use loss-gradient-weighted eNTK
position or activation and per-sample gradient spectra (Nikolaou et al.,
2026; Liu, Paquette, and Sous, 2026).

\subsection{3.6 From Theoretical Objects to Falsifiable
Predictions}\label{from-theoretical-objects-to-falsifiable-predictions}

At one matched budget, the relevant prediction is comparative: does a
prespecified geometry measurement identify which system has lower loss?
The completed prefix alone may fail. With target powers
\((.30,.25,.20,.15,.06,.04)\) and three acquired directions, the sets
\(\{ 1,2,6\}\) and \(\{ 1,3,4\}\) give

\[\left( J_{1},E_{1} \right) = (2,.41),\quad\quad\left( J_{2},E_{2} \right) = (1,.35).\]

The second system has a shorter prefix but a smaller residual because it
captures more energy beyond the first gap. A level prediction should
therefore use the full rank-resolved profile or a proxy validated on
held-out data.

Across budgets, the direct route predicts an interval
\(\lbrack\rho\gamma,\gamma\rbrack\), sharpened to (4) for power-law
target powers. Prefix adequacy selects the product endpoint. With a
common within-task tail, opposite acquisition-rate advantages across two
tasks predict the exponent reversal in Corollary 1. The fixed-kernel
route predicts the training-time rate from \(\beta\). These predictions
can be tested against finite residuals, fitted exponents, or both,
depending on the available resource range.

For stochastic systems, define the geometry and loss estimands at the
same distributional level. Shared-exponent resource rescaling supplies a
practical baseline. A geometry-based rate claim is challenged when its
conditions and measured separation hold but the loss exponents are
equivalent, or when their ordering runs opposite to the prediction.
Figure 2 summarizes the two exact routes and the corresponding tests.

\includegraphics[width=6.25in,height=4.03564in]{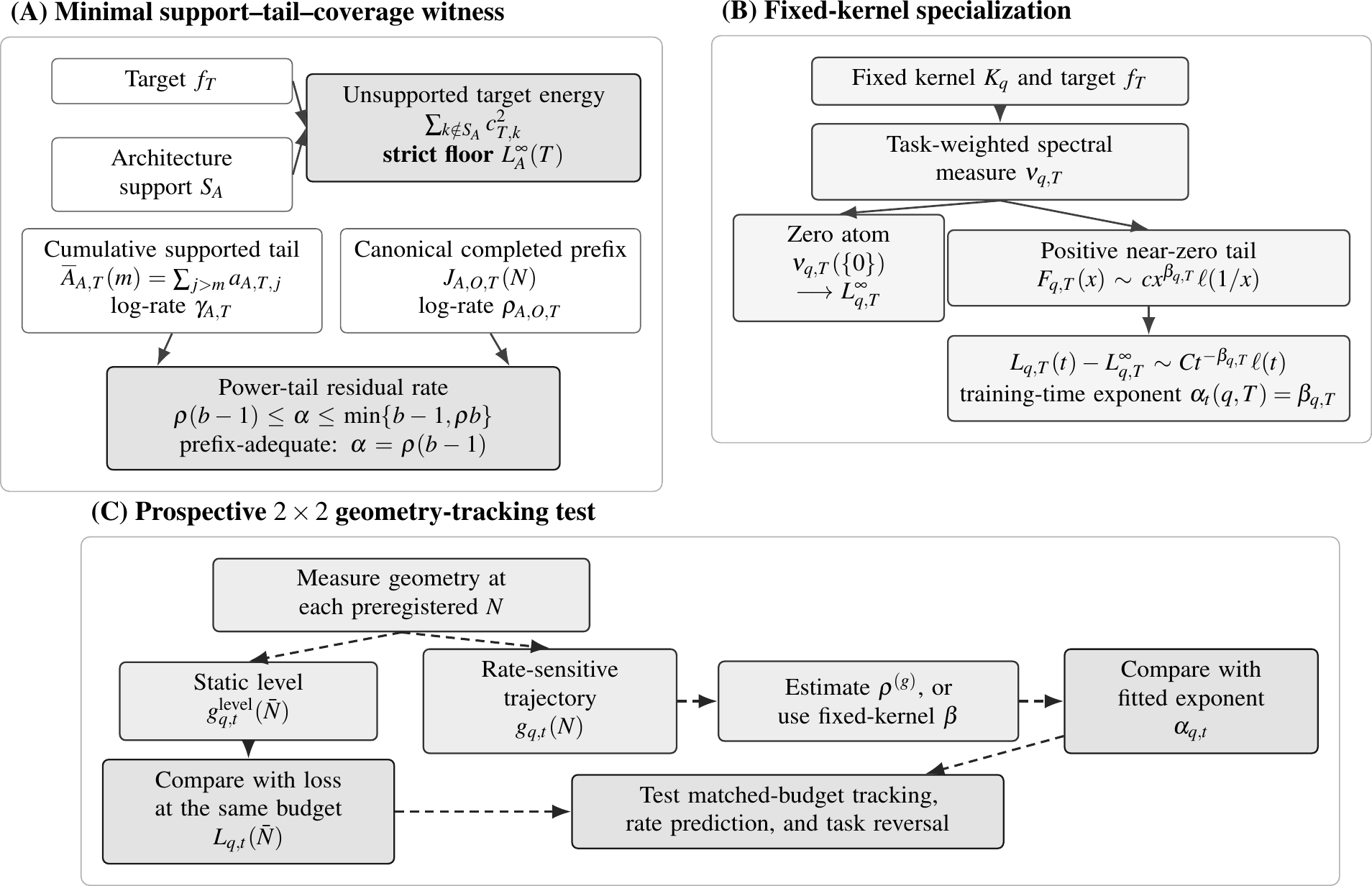}

\textbf{Figure 2. From support and acquisition to prospective
prediction.} Panel A gives the orthogonal route: unsupported energy sets
the floor, while the target tail and acquisition profile constrain the
residual rate. Panel B gives the fixed-kernel route, where task-weighted
mass near eigenvalue zero determines the training-time exponent. Panel C
separates a same-budget level test from a multiscale rate test. Dashed
arrows denote predictions evaluated on held-out data; the figure reports
no additional experiment.

\section{4. Evidence, Identification, and a Direct
Test}\label{evidence-identification-and-a-direct-test}

\subsection{4.1 From Acquired Modes to Observed Capability
Trajectories}\label{from-acquired-modes-to-observed-capability-trajectories}

The exact models track directions within one target. Public benchmark
releases usually track scores on different tasks across checkpoints. An
earlier task score does not reveal which modes were acquired or which
geometric rate produced the change. The released-data audit identifies
comparison problems for a direct geometry-to-scaling experiment.

\subsection{4.2 Reanalyzing Released Emergence
Trajectories}\label{reanalyzing-released-emergence-trajectories}

Even a smooth power-law curve mixes several determinants of emergence
timing. Let \(x\) be a monotone scale coordinate and suppose
\(L_{t}(x) = L_{t}^{\infty} + C_{t}x^{- \alpha_{t}}\) in a fitted
regime. For \(C_{t} > 0\) and threshold
\(\varepsilon_{t} > L_{t}^{\infty}\),

\[x_{t}(A,O) = \left( \frac{C_{t}(A,O)}{\varepsilon_{t} - L_{t}^{\infty}(A,O)} \right)^{1/\alpha_{t}(A,O)}.\]

The crossing order can change because the exponent, coefficient, or
floor changes. Prerequisite-linked tasks add a second source of order:
under commensurate evaluations, success on a composite may entail
competence on a component. An observed sequence can therefore mix
learning-system effects, task dependencies, and measurement choices.
This is the identification problem behind our reanalysis of the
implicit-curriculum trajectories reported by Liu et al.~(2026).

The released suite provides four trajectories from three open-weight
model families, with scale, training recipe, checkpoint structure, and
evaluation varying jointly. We retain five accuracy thresholds and two
threshold-free summaries. AUC measures normalized trajectory area over
checkpoint index; max-slope records the largest smoothed positive step.
Pairwise agreement is the fraction of valid model pairs that preserve
the same strict task order. Appendix A.4 gives the exact definitions and
archived implementation.

The primary universe contains 29 tasks and 46 prerequisite edges; a
registry-native universe provides a robustness comparison. The unmatched
A--B benchmark compares prerequisite-linked elemental--composite pairs
with unrelated elemental pairs, but it also changes task level. The main
comparison therefore holds each composite \(c\) fixed. For its linked
elemental pairs \(A(c)\) and unlinked elemental pairs \(C_{1}(c)\),
define

\[\Delta(c) = {\overline{S}}_{A(c)} - {\overline{S}}_{C_{1}(c)},\quad\quad\overline{\Delta} = \frac{1}{\left| \mathcal{C} \right|}\sum_{c\mathcal{\in C}}^{}\Delta(c).\]

Each qualifying composite receives equal weight. Elemental identity,
difficulty, format, and exposure remain uncontrolled, so the matched
contrast is descriptive.

\textbf{Table 2. Within-composite matched contrasts.}
\(\overline{\Delta}\) is the composite-equal-weight contrast;
\(r_{perm}\) is the uncalibrated within-composite permutation-reference
tail area in the prespecified positive direction. Ranges are the
2.5--97.5\% composite-resampling percentiles. Appendix A.5 explains
their status.

\begin{longtable}[]{@{}
  >{\raggedright\arraybackslash}p{(\columnwidth - 10\tabcolsep) * \real{0.1667}}
  >{\raggedright\arraybackslash}p{(\columnwidth - 10\tabcolsep) * \real{0.1667}}
  >{\raggedright\arraybackslash}p{(\columnwidth - 10\tabcolsep) * \real{0.1667}}
  >{\raggedright\arraybackslash}p{(\columnwidth - 10\tabcolsep) * \real{0.1667}}
  >{\raggedright\arraybackslash}p{(\columnwidth - 10\tabcolsep) * \real{0.1667}}
  >{\raggedright\arraybackslash}p{(\columnwidth - 10\tabcolsep) * \real{0.1667}}@{}}
\toprule\noalign{}
\begin{minipage}[b]{\linewidth}\raggedright
Measure
\end{minipage} & \begin{minipage}[b]{\linewidth}\raggedright
Universe
\end{minipage} & \begin{minipage}[b]{\linewidth}\raggedright
\[n_{c}\]
\end{minipage} & \begin{minipage}[b]{\linewidth}\raggedright
\[\overline{\Delta}\]
\end{minipage} & \begin{minipage}[b]{\linewidth}\raggedright
\[r_{perm}\]
\end{minipage} & \begin{minipage}[b]{\linewidth}\raggedright
Percentile range
\end{minipage} \\
\midrule\noalign{}
\endhead
\bottomrule\noalign{}
\endlastfoot
max-slope & Primary & 17 & −0.068 & 0.925 & (−0.175, +0.033) \\
max-slope & Registry & 12 & −0.103 & 1.000 & (−0.228, −0.011) \\
AUC & Primary & 20 & +0.029 & 0.176 & (−0.035, +0.093) \\
AUC & Registry & 17 & +0.084 & 0.020 & (−0.031, +0.209) \\
\end{longtable}

Matching substantially weakens the unmatched pattern. In the primary
universe, the A--B benchmark is positive for max-slope, AUC, and all
five thresholds. Once the composite is held fixed, max-slope becomes
negative in both universes, while AUC remains positive but small; both
AUC resampling ranges span zero. The apparent curriculum is therefore
sensitive to the comparison and the measure.

The sensitivity analysis gives AUC the stronger descriptive role. Its
matched contrast remains positive under every model deletion and after
removing six operationally degenerate composites, at +0.040 in the
primary universe and +0.044 in the registry universe. Max-slope changes
with checkpoint structure, weighting, and model deletion and remains
exploratory. Because the released trajectories cannot identify the
proposed accessibility mechanism, a direct test must isolate the
coupling intervention, control or deliberately cross task structure,
measure geometry independently of task scores, and replicate training
across seeds.

\subsection{4.3 A Staged Multiscale
Test}\label{a-staged-multiscale-test}

\textbf{Stage 1: test the exact bound where the objects are observable.}
Use a small system with a known task basis, target powers, acquired
ranks, and several budgets. The primary outcomes are the finite bound in
(1), the distinction between prefix-limited and dispersed acquisition,
and the rank-window criterion in Lemma 1. This stage tests the
mathematics in a system where the full representation is observable.

\textbf{Stage 2: test whether geometry predicts a feature-learning
system.} Cross at least two architecture--optimization couplings with
two task families chosen for an expected alignment reversal. Within each
task instance, hold examples, objective, evaluation, data exposure, and
seed distribution common at every budget outside the declared
intervention. Fix one primary resource axis and state how the remaining
resources vary along it. Parameter-matched and compute-matched designs
estimate different comparisons.

At successive budgets, measure task-conditioned geometry before fitting
the loss curves. Freeze the probe population, target, layer, capacity,
normalization, and rank convention. Use separate data for model
training, probe fitting, probe evaluation, and final task evaluation.
Measurements from lower budgets then predict held-out budgets and
independently sampled task instances.

At a matched budget \(\bar{N}\), orient the geometry scalar so that
larger values predict better performance. The level hypothesis is

\[sign\left\lbrack L_{q_{2},t}\left( \bar{N} \right) - L_{q_{1},t}\left( \bar{N} \right) \right\rbrack = sign\left\lbrack g_{q_{1},t}^{level}\left( \bar{N} \right) - g_{q_{2},t}^{level}\left( \bar{N} \right) \right\rbrack.\]

This is kept separate from the rate test because a system can have lower
loss at one budget and a shallower curve.

For a rank-interpretable trajectory, estimate
\(J_{q,t}^{(g)}(N) \asymp N^{\rho_{q,t}^{(g)}}\). With a common
within-task tail rate \(\gamma_{t}\) and prefix adequacy,

\[\alpha_{q,t}^{pred} = \rho_{q,t}^{(g)}\gamma_{t}.\]

The primary comparative prediction is a sign match between the
within-task acquisition-rate contrast and the within-task loss-exponent
contrast. A coupling-by-task reversal requires opposite signs on the two
tasks, with joint uncertainty from all four curves; a nonzero
difference-in-differences with agreeing within-task signs does not
qualify.

A fixed-kernel study uses \(\beta_{q,t}\) as the predictor. A
rank-resolved study with substantial dispersed acquisition uses the
applicable interval from Proposition 2. These branches are chosen from
the measured object before the exponent results are examined. Total-loss
predictions also account for coefficient error and the physical resource
clock in Appendix B.6.

Evaluate held-out predictions against a resource-only shared-exponent
rescaling model, coupling-specific exponents, and prespecified broken or
other non-power-law alternatives. Carry the joint uncertainty in floor,
coefficient, and exponent through every contrast. Pilot runs estimate
reliability and cost; the confirmatory design then fixes sample size and
meaningful effect or equivalence margins. Seeds replicate training,
while independent task instances support task-family claims. Appendix
C.2 gives the implementation and inference details.

\section{5. Implications and Research
Priorities}\label{implications-and-research-priorities}

\subsection{5.1 Interpreting a Scaling
Wall}\label{interpreting-a-scaling-wall}

Under Coupled Scaling, a wall is specific to a task, coupling, resource
axis, and budget range and may reflect delayed acquisition within
eventual support, persistent coefficient error, a temporary data or
optimization bottleneck, or target energy outside support. These
mechanisms can look similar over a short window.

Different interventions separate them. Longer training that restores the
previous slope supports a transient bottleneck. A structurally different
coupling that improves the slope or attainable loss on matched data
motivates a search for changed task-relevant geometry. Convergence of
well-chosen couplings toward the same floor increases the plausibility
of irreducible noise or task-specification limits, although a credible
floor requires a sufficiently long resource range.

Across model generations, this suggests a staircase conjecture:
structural changes open additional task-relevant regions, and subsequent
scaling develops those regions. Establishing the pattern would require
matched interventions and geometry measurements across the transition.

\subsection{5.2 Task-Conditional Evaluation and Model
Selection}\label{task-conditional-evaluation-and-model-selection}

When system advantages vary across tasks, an aggregate ranking depends
on the task mixture used to construct it. Prompts, scoring rules, and
sampling weights all change the effective evaluation distribution. Model
selection should therefore be tied to the intended deployment
distribution, and a benchmark rank should be read as a summary of the
distribution on which it was measured.

The released-data audit gives a concrete reason to preserve
component-level trajectories in compositional evaluation. Endpoint
scores and emergence summaries mix prerequisite structure with task
level, format, exposure, and operational degeneracy. Component
trajectories and format-aligned comparisons make those contributions
easier to locate.

\subsection{5.3 The Next Empirical Step}\label{the-next-empirical-step}

The highest-information next step is the Stage 1 experiment: a small
system with directly observable target ranks and acquired directions. It
can test the first-missing-rank bound, measure off-prefix gain, and
evaluate the rank-window condition at low cost. A successful direct test
would then justify the more expensive network study in which
lower-budget geometry predicts held-out learning curves. Recurrent or
looped architectures offer one candidate setting for tracking expert
routing, residual updates, and attention across budgets, subject to
access to training and checkpoint artifacts (Wang et al., 2026).

The corresponding theoretical extension would derive support,
acquisition distributions, and resource clocks from architecture and
stochastic optimization while quantifying coefficient error. Independent
geometry measurement and predictive testing provide the empirical
constraints needed for that extension.

\subsection{5.4 Conclusion}\label{conclusion}

Coupled Scaling explains learning-system comparisons through a
task-conditioned, budget-relative relation between support and
acquisition. In the orthogonal model, unsupported target energy fixes
the asymptotic floor. The first missing rank, the completed prefix, and
the acquired positive-target count bound the finite-budget residual and
yield the sharp power-tail interval
\(\left\lbrack \rho(b - 1),\min\{ b - 1,\rho b\} \right\rbrack\). The
rank-window condition identifies when the interpretable product rate
\(\rho(b - 1)\) applies, and common-tail comparisons convert cross-task
acquisition reversals into exponent reversals.

The fixed-kernel specialization reaches the same comparative question
through the task-weighted spectral tail. The empirical contribution is a
testable sequence: verify the exact bound where ranks are observable,
then ask whether independently measured geometry improves held-out
scaling prediction and anticipates which coupling benefits more from
additional resources on each task. That sequence turns representational
accessibility into a falsifiable account of when scaling rates should
agree and when they should change.

\section{Appendix A. Reanalysis
Protocol}\label{appendix-a.-reanalysis-protocol}

This appendix documents the data construction, ordering measures, and
statistical procedures for §4.2. The archived scripts and
machine-readable outputs are described in A.11. The result tables retain
the canonical estimates and report subsequent sensitivity analyses
separately.

\subsection{A.1 Data Source and
Models}\label{a.1-data-source-and-models}

The data are from Liu et al.~(2026), released at
https://github.com/KaiserWhoLearns/ElementalTask. The analysis uses
commit \texttt{fc08c318}, which contained no license file; a project
author subsequently confirmed that the code and data are MIT-licensed.
The scripts read the source data in place. Released accuracy
trajectories cover Pythia-6.9B, Amber-7B, OLMo2-1B, and OLMo2-7B, with
early-training checkpoint releases for the OLMo2 models. These are four
runs from three open-weight model families. Model family, size, corpus
or training recipe, data order, checkpoint structure, and training
randomness vary jointly; the two OLMo2 models use the same named OLMo-2
mixtures. Sixty-two tasks have trajectories for all four models. The
reanalysis uses these public trajectories, a subset of the broader model
collection described in the source paper.

\subsection{A.2 Two Task Universes}\label{a.2-two-task-universes}

The prerequisite relation \(t_{i} \prec t_{j}\) means that \(t_{i}\) is
a compositional component of \(t_{j}\). It is constructed from two
sources in the release. The \textbf{primary universe} follows the
authors' operation-to-component map in
\texttt{predict\_compositional\_from\_components.py}, including
\texttt{reverse} mapped to \texttt{token\_reversal}. It includes every
measured \texttt{compositional\_*} task whose operation chain parses
under that map and whose components are measured for all four models.
The result is 29 tasks: nine elementals and 20 composites, with 46
prerequisite edges and 406 task pairs. This construction includes every
measured composite that can be resolved into measured components under
the authors' mapping.

The \textbf{registry-native universe} follows the dataset metadata
directly: the \texttt{operations} columns of
\path{dataset/compositional*.csv} and the elementals in
\path{dataset/simple.csv}, retaining tasks measured by all four
models. It contains 27 tasks, comprising ten elementals and 17
composites, with 27 edges and 351 pairs. There is no registry elemental
for \texttt{reverse}, so reverse-containing composites connect only
through their other components. Country-to-capital and
country-to-currency tasks are isolated because the registered composites
do not use them. This universe is the robustness comparison.

Membership also differs where the release lacks a measured component.
The \texttt{last\_letter} entry exists in the registry but has no public
trajectory. Accordingly, \texttt{lower\_last} and \texttt{upper\_last}
each retain only a case-map edge in the registry universe and are absent
from the primary universe, which requires every component to be
measured. Conversely, parsable three-operation chains such as
\texttt{lower\_reverse\_first} may be absent from registry files.
Operational degeneracy is evaluated from each task's definition;
differences between the two exclusion lists follow from these membership
differences.

\subsection{A.3 Pair Categories and Operational
Degeneracy}\label{a.3-pair-categories-and-operational-degeneracy}

Pairs are classified by their position in the task DAG, independently of
performance and in the following order. Category A contains
prerequisite-linked pairs. Category B contains two elementals with no
dependency. Category C contains pairs at different compositional levels
with no prerequisite relation. For same-level composites, D contains
pairs with a shared component and E pairs with disjoint components. The
derived subtype \(C_{1}\) contains unlinked elemental--composite pairs:
134 in the primary universe and 143 in the registry universe. For each
composite \(c\), its elemental pairings partition into \(A(c)\) and
\(C_{1}(c)\). The primary category counts are A 46, B 36, C 218, D 59,
and E 47; the registry counts are 27, 45, 173, 48, and 58, respectively.
The main contrast holds \(c\) fixed and compares A with \(C_{1}\); A--B
and A--all-incomparable remain benchmarks.

Each universe contains six operationally degenerate composites. A
declared operation is degenerate when deleting it leaves the
input--output mapping unchanged for every item in that task's pool. Four
cases use single-character inputs, on which reversal and first- or
last-character extraction are identities. The primary cases are
\texttt{lower\_first}, \texttt{lower\_reverse}, and
\texttt{lower\_reverse\_first}, all reducing to lowercase, and
\texttt{upper\_reverse\_first}, reducing to uppercase. The registry
cases are \texttt{lower\_first}, \texttt{lower\_last}, and
\texttt{lower\_reverse}, reducing to lowercase, and
\texttt{upper\_last}, reducing to uppercase. Two further cases are
common to both universes: \texttt{plural\_lower} coincides with
\texttt{singular\_to\_plural}, and \texttt{translate\_eng\_fr\_lower}
with \texttt{translate\_eng\_fr}, because their outputs are already
lowercase. The corresponding French-to-English and Spanish-to-English
composites contain uppercase outputs and are not degenerate.

These tasks account for 14 of the 46 primary prerequisite edges and nine
of the 27 registry edges. Those edges encode operational identities.
Under the pair-validity rule, identical trajectories are tie-excluded,
but resampled demonstrations can separate them and produce a
noise-driven strict order. Exclusion is therefore performed at task
level using the definition-based rule above, independently of
performance. A.6c reports the complete excluded-task battery. Main-text
results include these tasks, with exclusions reported explicitly.

\subsection{A.4 Emergence Measures and Ordering
Agreement}\label{a.4-emergence-measures-and-ordering-agreement}

For model \(m\) and task \(t\), let the cleaned accuracy sequence be
\(a_{0},\ldots,a_{n - 1}\). Checkpoint identifiers are parsed
numerically and release rows such as \texttt{main}, which duplicate
final-model results, are removed. Threshold timing is the first index
with \(a_{i} \geq \theta\), for \(\theta \in \{ 0.3,0.4,0.5,0.6,0.7\}\).
It is undefined if the trajectory never reaches the threshold.

The archived max-slope implementation first applies a two-point moving
average,
\({\widetilde{a}}_{i} = \left( a_{i} + a_{i\text{+}1} \right)/2\), then
returns the earliest index maximizing
\({\widetilde{a}}_{i\text{+}1} - {\widetilde{a}}_{i}\). It is undefined
for fewer than four observations or when no finite positive improvement
exists. Checkpoint spacing is unweighted. The AUC implementation is the
trapezoidal integral over checkpoint index, normalized by \(n - 1\); it
uses its negative as the ordering score so that a larger area is ordered
earlier. These definitions follow \texttt{em\_maxslope} and
\texttt{em\_negauc} in the archived canonical script.\footnote{Canonical
  implementation: \texttt{reanalysis/scripts/reanalysis\_canonical.py},
  tag \texttt{v1.0-reanalysis}, functions \texttt{em\_maxslope},
  \texttt{em\_negauc}, and \texttt{frame}; blob SHA
  \texttt{1f2556b80b14a306747951b07f3b2c4f3e35504a}. See
  \url{https://github.com/quintonvina/coupled-scaling/blob/v1.0-reanalysis/reanalysis/scripts/reanalysis_canonical.py}.
  The definitions here document the archived calculations; the reported
  numerical results are retained.}

For a chosen measure, let \(s_{m}(i,j)\) be \(+ 1\) when task \(i\) is
ordered before \(j\) in model \(m\), and \(- 1\) for the reverse. Ties
and undefined scores are invalid. If \(\mathcal{V}_{ij}\) is the set of
model pairs for which both strict orders are valid, the agreement score
computed by \texttt{frame} is

\[S(i,j) = \frac{1}{\left| \mathcal{V}_{ij} \right|}\sum_{\left( m,m^{'} \right) \in \mathcal{V}_{ij}}^{}\mathbf{1}\{ s_{m}(i,j) = s_{m^{'}}(i,j)\}.\]

A task pair enters the analysis only when
\(\left| \mathcal{V}_{ij} \right| > 0\). Four models allow at most six
model-pair comparisons, but valid counts vary across task pairs and
measures. Category averages give equal weight to task pairs; the
within-composite estimate subsequently gives equal weight to composites.
A.12 examines the effect of changing these weights.

Prompt construction differs by task level. Elemental tasks use a fixed
demonstration set across checkpoints; compositional prompts are redrawn
at each evaluation by an unseeded generator using
\texttt{random.shuffle} in \texttt{compositional\_task.py}. The
within-composite contrast holds this composite-side sampling component
common to both groups. Degenerate component pairings remain vulnerable
because their entire strict order can come from sampling noise. A
fixed-prompt protocol in the prospective design removes this source of
repeated-evaluation variation.

\subsection{A.5 Statistical
Procedures}\label{a.5-statistical-procedures}

The primary declared comparison is the within-composite A--\(C_{1}\)
contrast. The A--B and A--all-incomparable benchmarks use 10,000 label
permutations within the compared categories. Their task-resampling
ranges use 2,000 draws of the task set with replacement. A replicate
keeps a pair if both tasks appear in the drawn set and computes the
statistic on that induced subgraph, discarding repeat-draw multiplicity.
Shared tasks induce dependence between pairs, and the interval
conservativeness of this procedure is undetermined.
Permutation-reference tail areas are consequently reported as
descriptive quantities; the multiplicity-weighted sensitivity in A.12
assesses an alternative resampling scheme.

Task pairs reuse both tasks and model trajectories, so pair observations
are dependent. The archived one-sided label-permutation values are
reported as descriptive reference values. Its 2,000 task resamples are
reduced to the unique included task set before the pair statistics are
recomputed; the resulting ranges are subset-perturbation sensitivity
intervals. The public-data analysis is diagnostic; confirmatory evidence
requires the direct test in Section 4.3.

For the matched contrast, each composite has \(k(c)\) valid component
pairs and \(m(c)\) valid non-component pairs. A within-composite
permutation reassigns which \(k(c)\) of the pooled pairs carry the
component label while keeping their \(S\) values fixed. The statistic is
the equal-composite mean \(\overline{\Delta}\). We use 10,000 draws from
an independent generator with seed 20260721. Main results report
\(r_{perm}\) in the \(\overline{\Delta} > 0\) direction documented
before implementation; the stated robustness results use the observed
direction. A two-sided reference area is twice the observed-direction
one-sided area, capped at one. The measures and positive direction were
documented before the matched implementation and its seeds were added,
but the analysis was not externally preregistered.

Calibration would require the relevant exchangeability assumptions;
exact randomization inference would require randomly assigned component
labels. In this observational suite, elemental identity, difficulty,
format, and prevalence across composites can covary with those labels.
The reference areas are therefore not calibrated significance
probabilities for a causal prerequisite effect.

Composite-resampling ranges use 2,000 multiplicity-preserving draws of
qualifying composites' \(\Delta(c)\) values, with seed 20260722. This
interval model treats composite contrasts as exchangeable. A
leave-one-elemental-out analysis reruns the matched procedure after
removing each elemental in turn to assess dependence arising from shared
components. The complete battery is reported without selection or
multiplicity correction, and its interpretation rests on effect size and
measure dependence.

\subsection{A.6a Benchmark Results}\label{a.6a-benchmark-results}

The following tables give the canonical category means, with valid
task-pair counts in parentheses. The primary table is displayed in two
parts to keep category values and contrasts readable; all entries retain
their original values. Reference areas are one-sided.

\textbf{Primary universe: category means.}

\begin{longtable}[]{@{}
  >{\raggedright\arraybackslash}p{(\columnwidth - 10\tabcolsep) * \real{0.1667}}
  >{\raggedright\arraybackslash}p{(\columnwidth - 10\tabcolsep) * \real{0.1667}}
  >{\raggedright\arraybackslash}p{(\columnwidth - 10\tabcolsep) * \real{0.1667}}
  >{\raggedright\arraybackslash}p{(\columnwidth - 10\tabcolsep) * \real{0.1667}}
  >{\raggedright\arraybackslash}p{(\columnwidth - 10\tabcolsep) * \real{0.1667}}
  >{\raggedright\arraybackslash}p{(\columnwidth - 10\tabcolsep) * \real{0.1667}}@{}}
\toprule\noalign{}
\begin{minipage}[b]{\linewidth}\raggedright
Measure
\end{minipage} & \begin{minipage}[b]{\linewidth}\raggedright
A
\end{minipage} & \begin{minipage}[b]{\linewidth}\raggedright
B
\end{minipage} & \begin{minipage}[b]{\linewidth}\raggedright
C
\end{minipage} & \begin{minipage}[b]{\linewidth}\raggedright
D
\end{minipage} & \begin{minipage}[b]{\linewidth}\raggedright
E
\end{minipage} \\
\midrule\noalign{}
\endhead
\bottomrule\noalign{}
\endlastfoot
\(\theta = 0.3\) & .941 (17) & .691 (27) & .858 (54) & .857 (7) & .556
(3) \\
\(\theta = 0.4\) & .818 (11) & .583 (24) & .863 (51) & 1.000 (8) & 1.000
(3) \\
\(\theta = 0.5\) & .833 (12) & .754 (19) & .865 (52) & .889 (9) & .333
(3) \\
\(\theta = 0.6\) & .909 (11) & .867 (20) & .939 (55) & 1.000 (8) & 1.000
(2) \\
\(\theta = 0.7\) & .917 (12) & .841 (21) & .922 (60) & .917 (12) & 1.000
(3) \\
max-slope & .871 (31) & .593 (27) & .910 (108) & .852 (18) & 1.000
(19) \\
AUC & .906 (46) & .778 (36) & .890 (218) & .825 (59) & .894 (47) \\
\end{longtable}

\textbf{Primary universe: contrasts and resampling range.}

\begin{longtable}[]{@{}
  >{\raggedright\arraybackslash}p{(\columnwidth - 6\tabcolsep) * \real{0.2500}}
  >{\raggedright\arraybackslash}p{(\columnwidth - 6\tabcolsep) * \real{0.2500}}
  >{\raggedright\arraybackslash}p{(\columnwidth - 6\tabcolsep) * \real{0.2500}}
  >{\raggedright\arraybackslash}p{(\columnwidth - 6\tabcolsep) * \real{0.2500}}@{}}
\toprule\noalign{}
\begin{minipage}[b]{\linewidth}\raggedright
Measure
\end{minipage} & \begin{minipage}[b]{\linewidth}\raggedright
A--B gap (\(r_{perm}\))
\end{minipage} & \begin{minipage}[b]{\linewidth}\raggedright
A--all gap (\(r_{perm}\))
\end{minipage} & \begin{minipage}[b]{\linewidth}\raggedright
95\% range, A--B
\end{minipage} \\
\midrule\noalign{}
\endhead
\bottomrule\noalign{}
\endlastfoot
\(\theta = 0.3\) & .250 (.0079) & .143 (.0446) & (−.167, .579) \\
\(\theta = 0.4\) & .235 (.0715) & .016 (.4497) & (−.833, .667) \\
\(\theta = 0.5\) & .079 (.3402) & .010 (.4290) & (−.883, .500) \\
\(\theta = 0.6\) & .042 (.3856) & −.020 (.7275) & (−.500, .333) \\
\(\theta = 0.7\) & .075 (.3613) & .010 (.6100) & (−.500, .389) \\
max-slope & .278 (.0070) & .007 (.4992) & (−.085, .818) \\
AUC & .128 (.0176) & .037 (.1839) & (−.040, .322) \\
\end{longtable}

\textbf{Registry-native universe.}

\begin{longtable}[]{@{}
  >{\raggedright\arraybackslash}p{(\columnwidth - 12\tabcolsep) * \real{0.1429}}
  >{\raggedright\arraybackslash}p{(\columnwidth - 12\tabcolsep) * \real{0.1428}}
  >{\raggedright\arraybackslash}p{(\columnwidth - 12\tabcolsep) * \real{0.1428}}
  >{\raggedright\arraybackslash}p{(\columnwidth - 12\tabcolsep) * \real{0.1428}}
  >{\raggedright\arraybackslash}p{(\columnwidth - 12\tabcolsep) * \real{0.1428}}
  >{\raggedright\arraybackslash}p{(\columnwidth - 12\tabcolsep) * \real{0.1428}}
  >{\raggedright\arraybackslash}p{(\columnwidth - 12\tabcolsep) * \real{0.1428}}@{}}
\toprule\noalign{}
\begin{minipage}[b]{\linewidth}\raggedright
Measure
\end{minipage} & \begin{minipage}[b]{\linewidth}\raggedright
A
\end{minipage} & \begin{minipage}[b]{\linewidth}\raggedright
B
\end{minipage} & \begin{minipage}[b]{\linewidth}\raggedright
C
\end{minipage} & \begin{minipage}[b]{\linewidth}\raggedright
D
\end{minipage} & \begin{minipage}[b]{\linewidth}\raggedright
E
\end{minipage} & \begin{minipage}[b]{\linewidth}\raggedright
A--B gap (\(r_{perm}\))
\end{minipage} \\
\midrule\noalign{}
\endhead
\bottomrule\noalign{}
\endlastfoot
\(\theta = 0.3\) & 1.000 (6) & .793 (29) & .798 (28) & .857 (7) & --- &
.207 (.2316) \\
\(\theta = 0.4\) & .800 (5) & .829 (35) & .871 (31) & 1.000 (8) & --- &
−.029 (.7469) \\
\(\theta = 0.5\) & .833 (6) & .790 (35) & .882 (34) & 1.000 (8) & --- &
.043 (.4578) \\
\(\theta = 0.6\) & .833 (6) & .892 (37) & .902 (34) & 1.000 (7) & --- &
−.059 (.7425) \\
\(\theta = 0.7\) & .857 (7) & .839 (29) & .868 (38) & .923 (13) & 1.000
(2) & .018 (.5579) \\
max-slope & .784 (17) & .593 (27) & .781 (93) & .611 (12) & .818 (22) &
.192 (.1016) \\
AUC & .901 (27) & .759 (45) & .833 (173) & .792 (48) & .839 (58) & .142
(.0163) \\
\end{longtable}

The primary A--B contrast is positive under every measure, but all
task-resampling ranges include zero. The registry threshold results use
only five to seven prerequisite pairs and change sign at \(\theta = .4\)
and \(.6\). The lowest-threshold and AUC directions agree across
universes. Cells with \(n \leq 3\) are retained for completeness and
excluded from interpretation. Here \(n\) counts task pairs that
contribute at least one valid model comparison.

\subsection{A.6b Matched-Contrast
Results}\label{a.6b-matched-contrast-results}

\textbf{Table A.1. Within-composite matched results for all seven
measures.} Part (a) reports the equal-composite contrast, its
positive-direction reference area, and composite-resampling range. Part
(b) reports the pooled group means and pooled contrast over the same
qualifying composites. The registry threshold rows have only four
qualifying composites and are retained as background. The archived
\path{matched_contrast_results.json} also supplies per-composite
contrasts and group sizes, exclusions, and leave-one-elemental-out
results.

\textbf{(a) Equal-composite estimates.}

\begin{longtable}[]{@{}
  >{\raggedright\arraybackslash}p{(\columnwidth - 10\tabcolsep) * \real{0.1667}}
  >{\raggedright\arraybackslash}p{(\columnwidth - 10\tabcolsep) * \real{0.1667}}
  >{\raggedright\arraybackslash}p{(\columnwidth - 10\tabcolsep) * \real{0.1667}}
  >{\raggedright\arraybackslash}p{(\columnwidth - 10\tabcolsep) * \real{0.1667}}
  >{\raggedright\arraybackslash}p{(\columnwidth - 10\tabcolsep) * \real{0.1667}}
  >{\raggedright\arraybackslash}p{(\columnwidth - 10\tabcolsep) * \real{0.1667}}@{}}
\toprule\noalign{}
\begin{minipage}[b]{\linewidth}\raggedright
Measure
\end{minipage} & \begin{minipage}[b]{\linewidth}\raggedright
Universe
\end{minipage} & \begin{minipage}[b]{\linewidth}\raggedright
\[n_{c}\]
\end{minipage} & \begin{minipage}[b]{\linewidth}\raggedright
\[\overline{\Delta}\]
\end{minipage} & \begin{minipage}[b]{\linewidth}\raggedright
\[r_{perm}\]
\end{minipage} & \begin{minipage}[b]{\linewidth}\raggedright
95\% range
\end{minipage} \\
\midrule\noalign{}
\endhead
\bottomrule\noalign{}
\endlastfoot
\(\theta = 0.3\) & Primary & 9 & +0.086 & .2026 & (−0.111, +0.284) \\
\(\theta = 0.3\) & Registry & 4 & +0.202 & .2095 & (+0.042, +0.411) \\
\(\theta = 0.4\) & Primary & 9 & +0.034 & .4193 & (−0.167, +0.237) \\
\(\theta = 0.4\) & Registry & 4 & +0.035 & .4402 & (−0.243, +0.278) \\
\(\theta = 0.5\) & Primary & 9 & +0.052 & .2911 & (−0.086, +0.182) \\
\(\theta = 0.5\) & Registry & 4 & +0.007 & .5435 & (−0.260, +0.223) \\
\(\theta = 0.6\) & Primary & 9 & +0.018 & .4531 & (−0.093, +0.110) \\
\(\theta = 0.6\) & Registry & 4 & +0.001 & .6621 & (−0.257, +0.171) \\
\(\theta = 0.7\) & Primary & 9 & +0.015 & .4228 & (−0.096, +0.101) \\
\(\theta = 0.7\) & Registry & 4 & +0.058 & .3905 & (−0.221, +0.242) \\
max-slope & Primary & 17 & −0.068 & .9251 & (−0.175, +0.033) \\
max-slope & Registry & 12 & −0.103 & 1.000 & (−0.228, −0.011) \\
AUC & Primary & 20 & +0.029 & .1762 & (−0.035, +0.093) \\
AUC & Registry & 17 & +0.084 & .0198 & (−0.031, +0.209) \\
\end{longtable}

\textbf{(b) Pooled estimates.}

\begin{longtable}[]{@{}
  >{\raggedright\arraybackslash}p{(\columnwidth - 8\tabcolsep) * \real{0.2000}}
  >{\raggedright\arraybackslash}p{(\columnwidth - 8\tabcolsep) * \real{0.2000}}
  >{\raggedright\arraybackslash}p{(\columnwidth - 8\tabcolsep) * \real{0.2000}}
  >{\raggedright\arraybackslash}p{(\columnwidth - 8\tabcolsep) * \real{0.2000}}
  >{\raggedright\arraybackslash}p{(\columnwidth - 8\tabcolsep) * \real{0.2000}}@{}}
\toprule\noalign{}
\begin{minipage}[b]{\linewidth}\raggedright
Measure
\end{minipage} & \begin{minipage}[b]{\linewidth}\raggedright
Universe
\end{minipage} & \begin{minipage}[b]{\linewidth}\raggedright
\({\overline{S}}_{A}\) (\(n\))
\end{minipage} & \begin{minipage}[b]{\linewidth}\raggedright
\({\overline{S}}_{C_{1}}\) (\(n\))
\end{minipage} & \begin{minipage}[b]{\linewidth}\raggedright
Pooled gap (\(r_{perm}\))
\end{minipage} \\
\midrule\noalign{}
\endhead
\bottomrule\noalign{}
\endlastfoot
\(\theta = 0.3\) & Primary & 0.938 (16) & 0.793 (37) & +0.145 (.1739) \\
\(\theta = 0.3\) & Registry & 1.000 (6) & 0.754 (23) & +0.246 (.1059) \\
\(\theta = 0.4\) & Primary & 0.818 (11) & 0.800 (35) & +0.018 (.6678) \\
\(\theta = 0.4\) & Registry & 0.800 (5) & 0.840 (25) & −0.040 (.6424) \\
\(\theta = 0.5\) & Primary & 0.833 (12) & 0.806 (36) & +0.028 (.3434) \\
\(\theta = 0.5\) & Registry & 0.833 (6) & 0.857 (28) & −0.024 (.6755) \\
\(\theta = 0.6\) & Primary & 0.909 (11) & 0.905 (35) & +0.004 (.6558) \\
\(\theta = 0.6\) & Registry & 0.833 (6) & 0.881 (28) & −0.048 (.7787) \\
\(\theta = 0.7\) & Primary & 0.917 (12) & 0.904 (38) & +0.013 (.5636) \\
\(\theta = 0.7\) & Registry & 0.857 (7) & 0.815 (27) & +0.042 (.4394) \\
max-slope & Primary & 0.867 (30) & 0.931 (68) & −0.065 (.9721) \\
max-slope & Registry & 0.784 (17) & 0.859 (59) & −0.074 (1.000) \\
AUC & Primary & 0.906 (46) & 0.863 (134) & +0.043 (.1895) \\
AUC & Registry & 0.901 (27) & 0.814 (143) & +0.088 (.0619) \\
\end{longtable}

\subsection{A.6c Results Excluding Degenerate
Tasks}\label{a.6c-results-excluding-degenerate-tasks}

Applying the task-level rule in A.3 leaves 23 tasks, 32 edges, and 253
pairs in the primary universe, and 21 tasks, 18 edges, and 210 pairs in
the registry universe. The excluded battery uses an independent
generator with seed 20260723.

\textbf{Primary: category means after exclusion.}

\begin{longtable}[]{@{}
  >{\raggedright\arraybackslash}p{(\columnwidth - 10\tabcolsep) * \real{0.1667}}
  >{\raggedright\arraybackslash}p{(\columnwidth - 10\tabcolsep) * \real{0.1667}}
  >{\raggedright\arraybackslash}p{(\columnwidth - 10\tabcolsep) * \real{0.1667}}
  >{\raggedright\arraybackslash}p{(\columnwidth - 10\tabcolsep) * \real{0.1667}}
  >{\raggedright\arraybackslash}p{(\columnwidth - 10\tabcolsep) * \real{0.1667}}
  >{\raggedright\arraybackslash}p{(\columnwidth - 10\tabcolsep) * \real{0.1667}}@{}}
\toprule\noalign{}
\begin{minipage}[b]{\linewidth}\raggedright
Measure
\end{minipage} & \begin{minipage}[b]{\linewidth}\raggedright
A
\end{minipage} & \begin{minipage}[b]{\linewidth}\raggedright
B
\end{minipage} & \begin{minipage}[b]{\linewidth}\raggedright
C
\end{minipage} & \begin{minipage}[b]{\linewidth}\raggedright
D
\end{minipage} & \begin{minipage}[b]{\linewidth}\raggedright
E
\end{minipage} \\
\midrule\noalign{}
\endhead
\bottomrule\noalign{}
\endlastfoot
\(\theta = 0.3\) & 0.889 (9) & 0.691 (27) & 0.875 (24) & --- & 0.667
(2) \\
\(\theta = 0.4\) & 0.833 (6) & 0.583 (24) & 0.768 (23) & --- & 1.000
(2) \\
\(\theta = 0.5\) & 0.857 (7) & 0.754 (19) & 0.747 (25) & --- & 0.333
(2) \\
\(\theta = 0.6\) & 1.000 (6) & 0.867 (20) & 0.942 (23) & --- & 1.000
(2) \\
\(\theta = 0.7\) & 1.000 (6) & 0.841 (21) & 0.944 (24) & 1.000 (1) &
1.000 (2) \\
max-slope & 0.861 (24) & 0.593 (27) & 0.910 (74) & 0.926 (9) & 1.000
(13) \\
AUC & 0.948 (32) & 0.778 (36) & 0.918 (134) & 0.910 (26) & 0.953 (25) \\
\end{longtable}

\textbf{Primary: A--B contrast after exclusion.}

\begin{longtable}[]{@{}
  >{\raggedright\arraybackslash}p{(\columnwidth - 4\tabcolsep) * \real{0.2276}}
  >{\raggedright\arraybackslash}p{(\columnwidth - 4\tabcolsep) * \real{0.3354}}
  >{\raggedright\arraybackslash}p{(\columnwidth - 4\tabcolsep) * \real{0.4370}}@{}}
\toprule\noalign{}
\begin{minipage}[b]{\linewidth}\raggedright
Measure
\end{minipage} & \begin{minipage}[b]{\linewidth}\raggedright
A--B gap (\(r_{perm}\))
\end{minipage} & \begin{minipage}[b]{\linewidth}\raggedright
95\% resampling range
\end{minipage} \\
\midrule\noalign{}
\endhead
\bottomrule\noalign{}
\endlastfoot
\(\theta = 0.3\) & +0.198 (.0750) & (−0.417, +0.513) \\
\(\theta = 0.4\) & +0.250 (.0914) & (−0.593, +0.667) \\
\(\theta = 0.5\) & +0.103 (.2694) & (−0.833, +0.467) \\
\(\theta = 0.6\) & +0.133 (.2412) & (+0.000, +0.333) \\
\(\theta = 0.7\) & +0.159 (.2996) & (+0.000, +0.400) \\
max-slope & +0.269 (.0158) & (−0.085, +0.714) \\
AUC & +0.170 (.0023) & (+0.050, +0.333) \\
\end{longtable}

\textbf{Registry: category means after exclusion.}

\begin{longtable}[]{@{}
  >{\raggedright\arraybackslash}p{(\columnwidth - 10\tabcolsep) * \real{0.1667}}
  >{\raggedright\arraybackslash}p{(\columnwidth - 10\tabcolsep) * \real{0.1667}}
  >{\raggedright\arraybackslash}p{(\columnwidth - 10\tabcolsep) * \real{0.1667}}
  >{\raggedright\arraybackslash}p{(\columnwidth - 10\tabcolsep) * \real{0.1667}}
  >{\raggedright\arraybackslash}p{(\columnwidth - 10\tabcolsep) * \real{0.1667}}
  >{\raggedright\arraybackslash}p{(\columnwidth - 10\tabcolsep) * \real{0.1667}}@{}}
\toprule\noalign{}
\begin{minipage}[b]{\linewidth}\raggedright
Measure
\end{minipage} & \begin{minipage}[b]{\linewidth}\raggedright
A
\end{minipage} & \begin{minipage}[b]{\linewidth}\raggedright
B
\end{minipage} & \begin{minipage}[b]{\linewidth}\raggedright
C
\end{minipage} & \begin{minipage}[b]{\linewidth}\raggedright
D
\end{minipage} & \begin{minipage}[b]{\linewidth}\raggedright
E
\end{minipage} \\
\midrule\noalign{}
\endhead
\bottomrule\noalign{}
\endlastfoot
\(\theta = 0.3\) & 1.000 (3) & 0.793 (29) & 0.848 (11) & --- & --- \\
\(\theta = 0.4\) & 1.000 (2) & 0.829 (35) & 0.722 (12) & --- & --- \\
\(\theta = 0.5\) & 1.000 (3) & 0.790 (35) & 0.778 (15) & --- & --- \\
\(\theta = 0.6\) & 1.000 (3) & 0.892 (37) & 0.911 (15) & --- & --- \\
\(\theta = 0.7\) & 1.000 (3) & 0.839 (29) & 0.846 (13) & 1.000 (1) &
--- \\
max-slope & 0.810 (14) & 0.593 (27) & 0.891 (61) & 0.867 (5) & 1.000
(10) \\
AUC & 1.000 (18) & 0.759 (45) & 0.955 (110) & 0.870 (18) & 1.000 (19) \\
\end{longtable}

\textbf{Registry: A--B contrast after exclusion.}

\begin{longtable}[]{@{}
  >{\raggedright\arraybackslash}p{(\columnwidth - 4\tabcolsep) * \real{0.2276}}
  >{\raggedright\arraybackslash}p{(\columnwidth - 4\tabcolsep) * \real{0.3354}}
  >{\raggedright\arraybackslash}p{(\columnwidth - 4\tabcolsep) * \real{0.4370}}@{}}
\toprule\noalign{}
\begin{minipage}[b]{\linewidth}\raggedright
Measure
\end{minipage} & \begin{minipage}[b]{\linewidth}\raggedright
A--B gap (\(r_{perm}\))
\end{minipage} & \begin{minipage}[b]{\linewidth}\raggedright
95\% resampling range
\end{minipage} \\
\midrule\noalign{}
\endhead
\bottomrule\noalign{}
\endlastfoot
\(\theta = 0.3\) & +0.207 (.4517) & (+0.000, +0.429) \\
\(\theta = 0.4\) & +0.171 (.6574) & (+0.000, +0.361) \\
\(\theta = 0.5\) & +0.210 (.2818) & (+0.000, +0.424) \\
\(\theta = 0.6\) & +0.108 (.5684) & (+0.000, +0.278) \\
\(\theta = 0.7\) & +0.161 (.4356) & (+0.000, +0.333) \\
max-slope & +0.217 (.0840) & (−0.150, +0.561) \\
AUC & +0.241 (.0004) & (+0.100, +0.411) \\
\end{longtable}

Threshold rows become sparse after exclusion: at most nine prerequisite
pairs remain in a primary threshold row and at most three in a registry
row. Under AUC, the A--B contrast increases to 0.170 in the primary
universe and 0.241 in the registry universe, with respective resampling
ranges (0.050, 0.333) and (0.100, 0.411). The matched contrast is
smaller, as the next table shows. Its reference areas use the observed
direction.

\begin{longtable}[]{@{}
  >{\raggedright\arraybackslash}p{(\columnwidth - 10\tabcolsep) * \real{0.1667}}
  >{\raggedright\arraybackslash}p{(\columnwidth - 10\tabcolsep) * \real{0.1667}}
  >{\raggedright\arraybackslash}p{(\columnwidth - 10\tabcolsep) * \real{0.1667}}
  >{\raggedright\arraybackslash}p{(\columnwidth - 10\tabcolsep) * \real{0.1667}}
  >{\raggedright\arraybackslash}p{(\columnwidth - 10\tabcolsep) * \real{0.1667}}
  >{\raggedright\arraybackslash}p{(\columnwidth - 10\tabcolsep) * \real{0.1667}}@{}}
\toprule\noalign{}
\begin{minipage}[b]{\linewidth}\raggedright
Measure
\end{minipage} & \begin{minipage}[b]{\linewidth}\raggedright
Universe
\end{minipage} & \begin{minipage}[b]{\linewidth}\raggedright
\(n_{c}\) full → excluded
\end{minipage} & \begin{minipage}[b]{\linewidth}\raggedright
\(\overline{\Delta}\) full → excluded
\end{minipage} & \begin{minipage}[b]{\linewidth}\raggedright
\(r_{perm}\) observed
\end{minipage} & \begin{minipage}[b]{\linewidth}\raggedright
95\% range, excluded
\end{minipage} \\
\midrule\noalign{}
\endhead
\bottomrule\noalign{}
\endlastfoot
max-slope & Primary & 17 → 12 & −0.068 → −0.124 & .021 & (−0.261,
0.000) \\
max-slope & Registry & 12 → 9 & −0.103 → −0.130 & .063 & (−0.278,
−0.018) \\
AUC & Primary & 20 → 14 & +0.029 → +0.040 & .098 & (−0.002, +0.088) \\
AUC & Registry & 17 → 11 & +0.084 → +0.044 & .158 & (+0.000, +0.106) \\
\end{longtable}

\subsection{A.7 Prerequisite Direction and Category
Heterogeneity}\label{a.7-prerequisite-direction-and-category-heterogeneity}

Agreement between models and compliance with prerequisite direction are
different statistics. Counting ties as satisfying the coarse-grid
constraint, components emerge no later than their composites in 63--69\%
of defined model-edge comparisons in the primary universe and 74--87\%
in the registry universe. Restricting threshold comparisons to strict
orders lowers the primary rate to 25--28\%. The strict inversions
concentrate on extraction-style elementals whose answer formats are
stricter than their composites. For example, at one OLMo2 checkpoint,
\texttt{compositional\_first\_upper} has accuracy 0.81 while
\texttt{simple\_icl\_first\_letter} has accuracy 0.07. This
concentration links the apparent violation to task design.

The measure also changes which directions and pairs dominate. In primary
strict comparisons, the component is earlier in 25.3\% of max-slope
comparisons (87 cases) and 62.4\% of AUC comparisons (178 cases). The
registry figures are 45.5\% (44) and 77.7\% (103). Timing-sensitive
measures retain fewer strict, defined comparisons, increasing the weight
of affected tasks among the survivors. This heterogeneity motivates the
A--B benchmark, while the within-composite contrast remains the primary
comparison.

\subsection{A.8 Sensitivity to the Task
Universe}\label{a.8-sensitivity-to-the-task-universe}

The primary construction admits \texttt{token\_reversal} as the measured
realization of \texttt{reverse}; the registry construction does not. The
registry also includes isolated knowledge elementals. These choices
affect the number and composition of valid prerequisite comparisons. At
the lowest threshold and under AUC, the A--B direction agrees across
universes, while its magnitude and threshold-level stability differ.

\subsection{A.9 Threshold-Free
Robustness}\label{a.9-threshold-free-robustness}

Threshold-crossing measures can be sensitive to the evaluation metric
(Schaeffer et al., 2023), motivating both threshold-free summaries. The
A--B direction is positive under max-slope and AUC in both universes. In
the primary universe the gaps are 0.278 and 0.128, with reference areas
.0070 and .0176; the registry AUC gap is 0.142, with reference area
.0163. A.12 shows why the two measures receive different evidential
weight: AUC retains its direction under the investigated weighting and
completeness changes, whereas max-slope does not.

\subsection{A.10 Within-Family and Cross-Family
Agreement}\label{a.10-within-family-and-cross-family-agreement}

The analysis contains one within-family pair, OLMo2-1B and OLMo2-7B,
which also differs in model size. Under AUC, this pair agrees on
incomparable-task order more often than cross-family pairs do: 0.931
over 347 task pairs versus 0.855 over 357 in the primary universe, and
0.927 versus 0.794 in the registry universe. The direction is the same
for \(\theta \leq .5\), including 0.910 versus 0.763 at \(\theta = .3\),
and reverses for \(\theta \geq .6\) where few comparisons remain. The
comparison is descriptive and lacks replication for a family-level
inference.

\subsection{A.11 Code and
Reproducibility}\label{a.11-code-and-reproducibility}

The public artifact is \url{https://github.com/quintonvina/coupled-scaling},
tag \texttt{v1.0-reanalysis}. The \texttt{reanalysis/} directory
contains scripts, machine-readable release-of-record results,
fresh-environment verification logs, \path{requirements.txt},
\path{MANIFEST.sha256}, and the entry point \path{run_all.sh}. The
four scripts are under \path{reanalysis/scripts/}; paths can be
configured through relative defaults and environment-variable overrides.

\path{reanalysis_canonical.py} uses seed 20260707 and writes
\path{canonical_results.json}, producing the canonical battery in
A.6a. It runs on CPU in minutes on a laptop.
\path{matched_contrast_canonical.py} executes that script unmodified
and adds the matched analysis with independent generators: permutation
seed 20260721, resampling seed 20260722, and excluded-battery seed
20260723. It writes \path{matched_contrast_results.json} and
\path{a6c_excluded_battery.json}, including per-composite contrasts
and group sizes, pooled results, and leave-one-elemental-out results.
\path{degeneracy_audit.py} implements the task-definition checks in
A.3. \path{audit_sensitivity.py} uses seed 20260727 and writes
\path{audit_sensitivity_results.json} and
\path{pair_level_audit.csv} without overwriting canonical outputs.
Together these artifacts document the reported analyses for both task
universes.

\subsection{A.12 Subsequent Statistical Sensitivity
Analysis}\label{a.12-subsequent-statistical-sensitivity-analysis}

After the canonical, matched, and excluded batteries were fixed and
archived, an independently implemented audit reconstructed the pipeline
from the pinned source commit. A second implementation reproduced its
findings, and a fresh-clone run reproduced
\path{canonical_results.json} bit for bit. These checks form the
existing archived verification record. The sensitivity battery uses seed
20260727 and leaves the canonical values and seeds unchanged. Its
pair-level table records task universe, task identities, category,
per-model signs, valid-model count, comparison denominator, \(S\),
composite anchor, component status, and exclusion status.

\textbf{Comparison weights.} The canonical estimator weights task pairs
equally, regardless of how many model comparisons contribute to each
\(S\). Under max-slope, 20 of 31 primary prerequisite pairs and 19 of 27
primary elemental pairs rest on a single valid comparison; the registry
universe has no four-model-complete prerequisite or elemental pairs. The
primary A--B gap is +0.278 with equal task-pair weights, +0.017 with
denominator weights, and −0.133 among complete cases, the last using
five prerequisite and eight elemental pairs. Under AUC, 43 of 46
prerequisite pairs and all 36 elemental pairs use all four models, and
the corresponding estimates are +0.128, +0.142, and +0.145. This
difference in stability supports AUC as the descriptive benchmark.

\textbf{Sparse contrasts and model deletion.} The per-composite contrast
is exactly zero for 13 of 17 qualifying primary composites under
max-slope and 11 of 20 under AUC; registry counts are eight of 12 and
nine of 17. The matched mean thus depends on a small number of nonzero
contrasts. Removing OLMo2-1B changes primary max-slope from −0.068 to
+0.022 and the registry result to 0.000 on four qualifying composites.
AUC stays positive under every model deletion, ranging from +0.012 to
+0.068 in the primary universe and +0.068 to +0.105 in the registry
universe.

\textbf{Resampling multiplicity.} A sensitivity analysis retains
repeated draws by assigning pair weight \(w_{i}w_{j}\), where \(w_{i}\)
and \(w_{j}\) are task draw counts. With 20,000 draws and seed 20260727,
the primary AUC A--B range changes from (−0.040, 0.322) to (−0.069,
0.354). The excluded-task AUC A--B ranges retain positive lower
endpoints: (0.011, 0.357) in the primary universe and (0.087, 0.429) in
the registry universe. This supplements the induced-subgraph resampling
procedure.

\textbf{AUC integration axis.} Canonical AUC integrates over checkpoint
index. Amber and Pythia have approximately uniform training-step grids,
while OLMo2 checkpoints are denser early in training and run from step
150 to beyond \(10^{5}\). Integrating over parsed training step and
normalizing within model changes 3.0\% and 1.5\% of within-model
task-pair orders in OLMo2-1B and OLMo2-7B, respectively, and none in
Amber or Pythia. The primary A--B gap becomes +0.100 and the matched
mean +0.061 (\(n_{c} = 20\)); the registry values become +0.161 and
+0.121 (\(n_{c} = 17\)). Excluded-task matched means are +0.034 in both
universes, with \(n_{c} = 14\) and 11. No reported direction changes.
The step-weighted result remains a separately archived sensitivity;
throughout the paper AUC denotes the checkpoint-index integral.

Taken together, these checks leave max-slope exploratory. AUC provides
the more stable observational summary across weighting, completeness,
exclusions, model deletion, and integration-axis changes. Its matched
residual remains small and positive. The archived canonical declaration
and seeds remain separate from these later checks.

\section{Appendix B. Proofs and
Extensions}\label{appendix-b.-proofs-and-extensions}

The proofs keep the task, basis, and resource protocol fixed. They first
separate support from acquisition, then derive the finite bounds,
asymptotic rates, and attaining orders. The last subsection restores
coefficient error and converts the slot budget to a physical resource.

\subsection{B.1 Setup}\label{b.1-setup}

Let \(f^{\text{*}} \in L^{2}(D)\), with countable orthonormal basis
\(\{\phi_{k}\}\) and coefficients
\(c_{k} = \langle f^{\text{*}},\phi_{k}\rangle\). Then
\(f^{\text{*}} = \sum_{k}^{}c_{k}\phi_{k}\) and
\(\sum_{k}^{}c_{k}^{2} = \parallel f^{\text{*}} \parallel^{2} < \infty\).
Population loss is \(L(f) = \parallel f - f^{\text{*}} \parallel^{2}\).
Observation noise would add a constant and is omitted in the exact
construction. An architecture--optimization system is represented by a
protocol-asymptotic support \(S_{A}\) and a bijective priority order
\(\pi_{A,O}\mathbb{:N \rightarrow}S_{A}\). Its acquired set is
\(I_{A,O}(N) = \{\pi_{A,O}(r):1 \leq r \leq N\}\). A nested capacity
family is assumed to realize the corresponding exact-coefficient
approximations jointly:

\[{\widehat{f}}_{N} = \sum_{r = 1}^{N}c_{\pi_{A,O}(r)}\phi_{\pi_{A,O}(r)}.\]

Support is assigned to architecture in this specialization;
optimization, parameterization, and feature learning act through
acquisition order. This isolates acquisition error. Its motivation
includes coarse-to-fine resolution (Sharma and Kaplan, 2022),
resolution-limited spectral models (Bahri et al., 2024), and quanta
truncation (Michaud et al., 2023). Appendix B.6 reintroduces error on
acquired coefficients.

\subsection{B.2 Decomposition, Rate Bounds, and
Attainment}\label{b.2-decomposition-rate-bounds-and-attainment}

\textbf{Proof of Proposition 1.} The residual has disjoint orthogonal
components outside support and in the unacquired part of support. Hence

\[L\left( {\widehat{f}}_{N} \right) = \sum_{k \notin S_{A}}^{}c_{k}^{2} + \sum_{r > N}^{}c_{\pi_{A,O}(r)}^{2}.\]

For \(V_{A} = \overline{span}\{\phi_{k}:k \in S_{A}\}\), the first
component is \(\Pi_{V_{A}^{\bot}}f^{\text{*}}\), so its squared norm is
\(L_{A}^{\infty}(T)\). It vanishes exactly when every nonzero target
coefficient is supported. The priority order leaves it unchanged. Since
\(\pi_{A,O}\) exhausts the support and the target energy is summable,
the second term tends to zero. Thus the displayed floor is also the
ordinary loss limit in this model. \(▫\)

\textbf{Proposition 2: finite bounds.} Write \(J = J_{N}\) and
\(M = M_{N}\). Rank \(J + 1\) is missing. Among all \(M\)-element rank
sets that omit \(J + 1\), nonincreasing powers make
\(\{ 1,\ldots,M + 1\} \setminus \{ J + 1\}\) maximize acquired
energy. Thus

\[\sum_{j \in K_{N}}^{}a_{j} \leq \sum_{j = 1}^{M + 1}a_{j} - a_{J\text{+}1}.\]

Subtracting from total supported energy \(\overline{A}(0)\) gives the
lower bound in (1). The maximizing set has completed prefix \(J\), so
equality is attainable. All ranks through \(J\) have been acquired,
which gives \(E(N) \leq \overline{A}(J)\). Since \(M \leq N\) and the
tail is decreasing, (2) follows. It also recovers the best-\(N\) bound
because

\[a_{J\text{+}1} + \overline{A}(N + 1) \geq a_{N\text{+}1} + \overline{A}(N + 1) = \overline{A}(N).\]

\textbf{General-tail rates.} The assumed log-rates compose as

\[\frac{-\log \overline{A}\left( J_{N} \right)}{\log N} = \frac{-\log \overline{A}\left( J_{N} \right)}{\log J_{N}}\frac{\log J_{N}}{\log N} \rightarrow \rho\gamma.\]

Applying logarithms to
\(\overline{A}(N) \leq E(N) \leq \overline{A}\left( J_{N} \right)\)
proves (3). If \(\overline{A}\left( J_{N} \right)/E(N) = N^{o(1)}\), the
logarithm of this ratio divided by \(\log N\) tends to zero, and both
residual log-rates equal \(\rho\gamma\). The bounded-gain condition
implies this subpolynomial-ratio condition.

\textbf{Power-law tails.} The integral test gives
\(\overline{A}(m) \asymp m^{- (b - 1)}\). The upper residual bound has
log-rate \(\rho(b - 1)\), while the two nonnegative terms of the lower
bound imply

\[E(N) \geq N^{- \rho b + o(1)},\quad\quad E(N) \geq N^{- (b - 1) + o(1)}.\]

These inequalities give (4). If \(\log M_{N}/\log N \rightarrow \kappa\),
using (1) retains the acquired-count information and replaces the upper
endpoint by \(\min\{\kappa(b - 1),\rho b\}\). Under bounded gain and
\(J_{N} \asymp N^{\rho}\), comparing \(E(N)\) with
\(\overline{A}\left( J_{N} \right)\) yields the two-sided order
\(E(N) \asymp N^{- \rho(b - 1)}\).

\textbf{Diagnosing gain beyond the prefix.} For sufficiently large \(N\)
with \(J_{N} > 1\), define

\[\delta_{N} = \frac{\log\left\lbrack \overline{A}\left( J_{N} \right)/E(N) \right\rbrack}{\log N} \geq 0.\]

The identity

\[\frac{-\log E(N)}{\log N} = \frac{-\log \overline{A}\left( J_{N} \right)}{\log J_{N}}\frac{\log J_{N}}{\log N} + \delta_{N}\]

gives \(\alpha = \rho\gamma + \delta\) when
\(\delta_{N} \rightarrow \delta\). Equation (3) gives
\(0 \leq \delta \leq (1 - \rho)\gamma\); the power-tail refinement gives
\(0 \leq \delta \leq \min\{(1 - \rho)(b - 1),\rho\}\). This diagnostic
locates the extra rate contribution after loss is observed. The
rank-window criterion gives a structural condition for \(\delta = 0\).

\textbf{Direct construction of the full power-tail interval.} Fix one
support
\(S_{A} = \{ i_{1},i_{2},\ldots\} \cup \{ z_{1},z_{2},\ldots\}\), where
\(i_{j}\) has power \(a_{j} \asymp j^{- b}\) and each \(z_{j}\) has zero
target power. For a prescribed rate

\[\theta \in \left\lbrack \rho(b - 1),\min\{ b - 1,\rho b\} \right\rbrack,\quad\quad\kappa = \frac{\theta}{b - 1} \in \lbrack\rho,1\rbrack,\]

partition the positive-target ranks into the dyadic set
\(D = \{ 2^{\ell}:\ell \geq 1\}\) and let
\(H = \{ h_{1} < h_{2} < \cdots\}\) enumerate the remaining ranks. Let
\(c(x)\) be the least integer not smaller than \(x\), and set

\[\pi\left( 3c\left( 2^{\ell/\rho} \right) - 2 \right) = i_{2^{\ell}},\quad\quad\pi\left( 3c\left( m^{1/\kappa} \right) - 1 \right) = i_{h_{m}}.\]

Fill every remaining slot successively with \(z_{1},z_{2},\ldots\). The
two positive-target schedules occupy different residue classes modulo
\(3\) and are strictly increasing; the multiples of \(3\) leave
infinitely many zero-target slots. Thus \(\pi\) is a fixed bijection
whose prefixes form a nested acquisition path.

At budget \(N\), the next unacquired dyadic rank is of order
\(N^{\rho}\). The second schedule has acquired
\(m_{N} \asymp N^{\kappa}\) elements of \(H\), and \(h_{m} \sim m\), so
its next unacquired rank is of order \(N^{\kappa}\). Since
\(\kappa \geq \rho\),

\[J_{N} \asymp \min\{ N^{\rho},N^{\kappa}\} \asymp N^{\rho}.\]

The unacquired dyadic energy is a geometric tail. Removing the dyadic
ranks changes a power-law tail after rank \(x\) only by
\(O\left( x^{- b} \right)\), smaller than the \(x^{- (b - 1)}\) tail
order. Hence

\[E_{D}(N) \asymp N^{- \rho b},\quad\quad E_{H}(N) = \sum_{m > m_{N}}^{}a_{h_{m}} \asymp N^{- \kappa(b - 1)}.\]

Therefore

\[E(N) = E_{D}(N) + E_{H}(N) \asymp N^{- \rho b} + N^{- \kappa(b - 1)} \asymp N^{- \theta},\]

where the last step uses \(\kappa(b - 1) = \theta \leq \rho b\). This
constructs every rate in the claimed closed interval on the same
support, including both endpoints, and completes Proposition 2. \(▫\)

\textbf{Proof of Lemma 1.} For some \(c_{0} > 0\), all sufficiently
large ranks satisfy \(a_{j} \geq c_{0}j^{- b}\). Each unresolved rank in
the window of (5) has power at least
\(c_{0}\left( \lambda J_{N} \right)^{- b}\), and there are at least
\(cJ_{N}\) such ranks. Therefore

\[E(N) \geq cJ_{N}c_{0}\left( \lambda J_{N} \right)^{- b} = cc_{0}\lambda^{- b}J_{N}^{- (b - 1)}.\]

Combining this with
\(E(N) \leq \overline{A}\left( J_{N} \right) \asymp J_{N}^{- (b - 1)}\)
proves the claim. \(▫\)

\textbf{Finite-budget calculations.} Figure 1 evaluates
\(a_{j} = j^{- 1.6}\) at 60 logarithmically spaced integer budgets from
128 to 262,144. Every curve follows one fixed order; regressions use the
upper half of the grid. The finite-window slopes summarize these points,
while the rate labels follow from the proof. Independent
integer-energy enumerations were used as implementation
checks. The asymptotic claims do not depend on those numerical checks;
the public-data audit uses the separate artifact in A.11.

\textbf{Proof of Corollary 1.} Proposition 2 gives
\(\alpha_{q,t} = \rho_{q,t}\gamma_{t}\). Since \(\gamma_{t} > 0\),
multiplying by it preserves the within-task ordering of \(\rho\), so the
assumed reversal transfers to \(\alpha\). If
\(\alpha_{1} > \alpha_{2}\), then

\[\frac{\log\left\lbrack E_{1}(N)/E_{2}(N) \right\rbrack}{\log N} \rightarrow - \left( \alpha_{1} - \alpha_{2} \right) < 0.\]

Thus \(E_{1}(N)/E_{2}(N) \rightarrow 0\), establishing the eventual
residual-loss order. \(▫\)

The comparison concerns residual loss above the floor. Equal within-task
floors, or a negligible floor difference over the stated budget range,
extend the ordering to total loss.

\subsection{B.3 Connections to Learning-System
Mechanisms}\label{b.3-connections-to-learning-system-mechanisms}

The model assigns different roles to support and order. A support change
can alter the floor and the supported target tail; an order change can
alter finite-budget residuals while leaving the floor fixed. These are
controlled analogues for interpreting the system comparisons in §2. The
exponent differences in neural force fields motivate examining both
channels, but architecture-specific strict floors remain an empirical
target (Ngo and Ravanbakhsh, 2026). Copying bounds illustrate
task-specific architectural costs (Jelassi et al., 2024); mapping those
costs to this model's asymptotic support requires an explicit resource
protocol.

Within fixed support, feature learning is represented by
reprioritization. Under the product conditions, raising an initially
poor \(\rho_{init,T} < 1\) toward one improves the exponent; when
\(\rho_{init,T} = 1\), the endpoint is already reached. This parallels
the easy/hard target distinction of Bordelon et al.~(2025) on their time
and compute axes. Preconditioning can likewise be represented on the
order side, consistent with the fitted-exponent dependence studied by
Ramani and Jain (2026). Feature-learning networks may also change
support and coefficient error, so the richer description in B.6 is
needed for that extension. Superposition is a separate geometric
mechanism: the overlap contribution derived by Liu, Liu, and Gore (2025)
lies beyond this orthogonal acquisition calculation.

The data-side limit is recovered with full support and \(\rho = 1\),
giving a zero representational floor and \(\alpha = b_{T} - 1\).
Restoring observation noise leaves its irreducible constant. Under the
usage-frequency hypothesis of Michaud et al.~(2023), corpus statistics
alone determine the order, so systems sharing that order have
model-independent ordering in this idealization. The audit in §4.2
examines the comparison and measurement issues encountered when
investigating such ordering in public task trajectories.

\subsection{B.4 Relation to Solvable Scaling
Models}\label{b.4-relation-to-solvable-scaling-models}

Spectral theories derive learning curves from target alignment,
eigenspectra, and statistical regime (Maloney et al., 2022; Bordelon et
al., 2020; Canatar et al., 2021; Bahri et al., 2024). Bordelon et
al.~(2024) add rank-constrained dynamics, where a top-\(k^{\star}\)
truncation and target tail produce scaling laws. The frontier accounts
of Zou et al.~(2026) and Song et al.~(2026) provide the neighboring
tail--frontier construction discussed in §2.1. Proposition 2 retains
that tail calculation while allowing interleaved target ranks and
coupling-specific support, and determines its sharp power-tail rate
limits. Corollary 1 gives the cross-task comparison. In the fixed-kernel
specialization, the joint spectral measure keeps eigenvalue strength
together with target energy; asymptotically aligned rankings recover a
product form.

Proposition 3 uses the Abelian direction of the classical Karamata
Laplace--Stieltjes theorem. Suppress \(q,T\), remove the atom at zero,
and set \(s = 2t\). Stieltjes integration by parts gives

\[\int_{(0,\infty)}^{}e^{- s\lambda}\, dF(\lambda) = s\int_{0}^{\infty}e^{- s\lambda}F(\lambda)\, d\lambda = \int_{0}^{\infty}e^{- u}F(u/s)\, du.\]

After division by \(F(1/s)\), regular variation and Potter bounds
justify dominated convergence to
\(\int_{0}^{\infty}e^{- u}u^{\beta}\, du = \Gamma(\beta + 1)\). This
proves the stated loss asymptotic from the near-zero spectral tail. The
converse identification requires the corresponding Tauberian conditions.

\subsection{B.5 Rate Classes and Network
Extension}\label{b.5-rate-classes-and-network-extension}

The rate interval assumes a fixed basis, ranked target powers, and a
slot budget. Finite support, exponentially decaying tails, and
nonconvergent tail log-rates define other cases. The network extension
discussed in §5.3 would replace prescribed support and order with
quantities derived from the architecture and stochastic training
procedure. It would then track target tails, completed prefixes, and
off-prefix gain under a network-native resource map.

\subsection{B.6 Coefficient Error and Resource
Clocks}\label{b.6-coefficient-error-and-resource-clocks}

Allow an estimated coefficient \({\widehat{c}}_{k}\) on each acquired
direction, with \(I_{N} \subseteq S_{A}\) and
\({\widehat{f}}_{N} = \sum_{k \in I_{N}}^{}{\widehat{c}}_{k}\phi_{k}\).
Orthonormality gives three disjoint residual components:

\[L\left( {\widehat{f}}_{N} \right) = \underbrace{\sum_{k \notin S_{A}}^{}c_{k}^{2}}_{L_{\mathrm{supp}}} + \underbrace{\sum_{k \in S_{A} \setminus I_{N}}^{}c_{k}^{2}}_{E(N)} + \underbrace{\sum_{k \in I_{N}}^{}\left( {\widehat{c}}_{k} - c_{k} \right)^{2}}_{Q(N)}.\quad(6)\]

All cross terms vanish. The first term is unsupported energy; the second
is acquisition error; the third is error within the acquired subspace
and may contain bias, variance, and optimization error. If the latter
two vanish, the first is the limiting loss. A persistent coefficient
error can add a floor. More generally, if \(E(x) \asymp x^{- u}\) and
\(Q(x) \asymp x^{- v}\) are nonnegative, then
\(E(x) + Q(x) \asymp x^{- \min\{ u,v\}}\). Thus acquisition determines
the total excess-loss rate when the coefficient term is controlled, for
example by \(Q = O(E)\).

A physical resource axis introduces a further conversion. Suppose the
same support, target powers, and order acquire \(n(x) = x^{s + o(1)}\)
slots at resource level \(x\), where \(s > 0\), and
\(E(n) = n^{- \alpha + o(1)}\). Substitution gives

\[E\left( n(x) \right) = x^{- s\alpha + o(1)}.\]

Changing the target distribution, mixture, support, or order along \(x\)
changes both the underlying comparison and its clock. The resource
protocol records these choices so that the physical-resource exponent
has a defined interpretation.

\section{Appendix C. Evidence and Test
Specification}\label{appendix-c.-evidence-and-test-specification}

\subsection{C.1 Evidence Matrix}\label{c.1-evidence-matrix}

\textbf{Table C.1. Learning-system and resource-protocol comparisons.}
The columns separate the intervention and its controls from the reported
outcome and the question left for a direct test.

\begin{longtable}[]{@{}
  >{\raggedright\arraybackslash}p{(\columnwidth - 6\tabcolsep) * \real{0.2500}}
  >{\raggedright\arraybackslash}p{(\columnwidth - 6\tabcolsep) * \real{0.2500}}
  >{\raggedright\arraybackslash}p{(\columnwidth - 6\tabcolsep) * \real{0.2500}}
  >{\raggedright\arraybackslash}p{(\columnwidth - 6\tabcolsep) * \real{0.2500}}@{}}
\toprule\noalign{}
\begin{minipage}[b]{\linewidth}\raggedright
Study
\end{minipage} & \begin{minipage}[b]{\linewidth}\raggedright
Intervention and controls
\end{minipage} & \begin{minipage}[b]{\linewidth}\raggedright
Reported outcome
\end{minipage} & \begin{minipage}[b]{\linewidth}\raggedright
Remaining question
\end{minipage} \\
\midrule\noalign{}
\endhead
\bottomrule\noalign{}
\endlastfoot
Ngo \& Ravanbakhsh (2026) & Equivariant/non-equivariant force fields;
same data and task & Architecture-dependent parameter, data, and compute
exponents & Positive architecture-specific strict floors; accessibility
mechanism \\
Tay et al.~(2023) & Ten LM architectures; matched pretraining and
evaluation & Architecture-dependent curves and ranks that change with
scale & Task-conditioned geometry underlying the differences \\
Liu, Liu, \& Gore (2025) & Weight-decay-controlled superposition; fixed
toy architecture and feature distribution; four open LM families &
Geometric interference produces width scaling & Task-conditioned
tracking; data and training-time scaling \\
Bordelon et al.~(2025) & Feature-learning regime; controlled tasks and
models & Hard-target training-time and compute exponents change &
Parameter- and data-scaling dependence \\
Ramani \& Jain (2026) & Preconditioning; same random-feature model
within each spectral condition & Optimizer-dependent fitted model-size
exponents & Transfer beyond controlled lazy random features; strict
floors \\
Jha \& Reagen (2026) & AdamW, Muon, NorMuon, low-rank Dion; common GPT
family, FineWeb-Edu, and FFN-width grid; one seed per cell & Rank
exponents differ; AdamW--Dion difference persists at matched perplexity
& Geometry-to-performance prediction; seed robustness \\
Volkova et al.~(2026) & Optimizers; common family, corpus, objective,
and \(N,D\) grid within each architecture--dataset instance & Separate
fits ill-conditioned; constrained shared-exponent rescaling improves
stability and extrapolation & Independent exponent equivalence; geometry
and task interaction \\
Bansal et al.~(2022) & NMT architecture and data conditions; common
corpus and scale protocol within comparisons & Data exponents mostly
stable under tested architecture/task setup, filtering, and iid noise;
back-translation degrades them & Stability beyond tested interventions
and targets \\
Li et al.~(2026) & Domain repetition; fixed tokens per parameter;
separate fixed-domain-fraction comparison & Domain-dependent repetition
optima rise mildly with size & Geometry prediction under matched
exposure \\
Xiao et al.~(2025) & Historical model pipelines; no factorial separation
& Capability-density trend & Separate effects of data, architecture,
optimization, and evaluation \\
\end{longtable}

\subsection{C.2 Test Specification}\label{c.2-test-specification}

\textbf{Resource protocol.} Fix the primary axis and how parameters,
unique training data, cumulative presentations, compute, and training
time vary along it. Record mixture weights, sampling order, and
optimizer schedule at every budget, separating the declared intervention
from shared settings. Fixed-data and proportional-data paths estimate
different resource comparisons. The clock conversion in B.6 applies when
the underlying support, target powers, and acquisition order remain
fixed.

\textbf{Independent measurement and validation.} Freeze probe targets,
capacity, regularization, layers, normalization, rank conventions, and
measurement scales. Separate model-training, probe-fitting,
probe-held-out, and final task-evaluation data. Estimate geometry-side
predictors at lower budgets and assess them on held-out budgets and
independent task instances. Choose the predictive loss before comparison
and evaluate competing models on the same cases. Record every
pilot-driven design choice before confirmatory data collection.

\textbf{Theory-to-prediction mapping.} A product-law cell specifies a
rank-interpretable prefix trajectory and a common within-task
supported-tail rate. Full support in an ex ante task basis is one route
to a shared tail; independent tail estimation is another. Prefix
adequacy is established through the rank-window condition or a separate
bounded- or exponent-neutral-gain argument. Use the general interval
under cumulative-tail log-rate assumptions and the refined interval
under power-law target powers. A fixed-kernel cell uses the
task-weighted near-zero spectral tail under Proposition 3. Loss-derived
acceleration is retained as a diagnostic, and total-loss predictions
address coefficient error through (6).

\textbf{Competing curve models.} Compare shared-exponent resource
rescaling, coupling-specific exponents, and prespecified broken or other
non-power-law families by held-out prediction. On a single power-law
axis, constant resource rescaling and a free multiplicative coefficient
are observationally equivalent; choose a reference normalization to
identify the parameterization. Fit floor, coefficient, and exponent
jointly and retain their dependence in predictions and contrasts. When
the observed range does not resolve the asymptotic regime, evaluate the
theoretical curves at the observed finite budgets.

\textbf{Replication and uncertainty.} Preserve each seed--task-instance
trajectory as a unit in resampling or hierarchical modeling. Checkpoints
describe a curve; seeds replicate training; independent task instances
provide replication for task-family claims. Use pilot data to estimate
variance, measurement reliability, and cost, then set sample size
through a prespecified precision or power criterion and a scientifically
chosen meaningful-effect threshold. Randomization inference follows the
actual experimental assignment mechanism.

\textbf{Contrasts and equivalence.} The primary geometry comparison is
within task. A positive task-specific rescaling preserves its sign but
can change an unstandardized difference-in-differences, so cross-task
magnitudes require a common baseline or normalization. Exponent reversal
requires opposite signs for the two within-task contrasts, with joint
uncertainty. An equivalence result challenges a directional prediction
when geometric rate separation and the applicable theory conditions are
established, and the joint interval for \(d_{\alpha}(t)\) lies within
the chosen equivalence margin. Broad intervals or unresolved geometric
separation leave the comparison unclassified. Static proxy accuracy is
assessed separately.

\section{References}\label{references}

Aghajanyan, A., Gupta, S., and Zettlemoyer, L. (2021). Intrinsic
dimensionality explains the effectiveness of language model fine-tuning.
In \emph{Proceedings of the 59th Annual Meeting of the Association for
Computational Linguistics and the 11th International Joint Conference on
Natural Language Processing (Volume 1: Long Papers)}, pages 7319--7328.

Ansuini, A., Laio, A., Macke, J. H., and Zoccolan, D. (2019). Intrinsic
dimension of data representations in deep neural networks. In
\emph{Advances in Neural Information Processing Systems 32 (NeurIPS)},
pages 6111--6122.

Arora, S., Eyuboglu, S., Timalsina, A., Johnson, I., Poli, M., Zou, J.,
Rudra, A., and Ré, C. (2024). Zoology: Measuring and improving recall in
efficient language models. In \emph{The Twelfth International Conference
on Learning Representations (ICLR)}.

Bahri, Y., Dyer, E., Kaplan, J., Lee, J., and Sharma, U. (2024).
Explaining neural scaling laws. \emph{Proceedings of the National
Academy of Sciences}, 121(27):e2311878121.

Bansal, Y., Ghorbani, B., Garg, A., Zhang, B., Cherry, C., Neyshabur,
B., and Firat, O. (2022). Data scaling laws in NMT: The effect of noise
and architecture. In \emph{Proceedings of the 39th International
Conference on Machine Learning}, PMLR 162:1466--1482.

Bingham, N. H., Goldie, C. M., and Teugels, J. L. (1987). \emph{Regular
Variation}. Cambridge University Press.

Bordelon, B., Atanasov, A., and Pehlevan, C. (2024). A dynamical model
of neural scaling laws. In \emph{Proceedings of the 41st International
Conference on Machine Learning (ICML)}, PMLR 235:4345--4382.

Bordelon, B., Atanasov, A., and Pehlevan, C. (2025). How feature
learning can improve neural scaling laws. \emph{Journal of Statistical
Mechanics: Theory and Experiment}, 2025(8):084002.

Bordelon, B., Canatar, A., and Pehlevan, C. (2020). Spectrum dependent
learning curves in kernel regression and wide neural networks. In
\emph{Proceedings of the 37th International Conference on Machine
Learning (ICML)}, PMLR 119:1024--1034.

Caballero, E., Gupta, K., Rish, I., and Krueger, D. (2023). Broken
neural scaling laws. In \emph{The Eleventh International Conference on
Learning Representations (ICLR)}.

Cagnetta, F., Favero, A., Sclocchi, A., and Wyart, M. (2025). Scaling
laws and representation learning in simple hierarchical languages:
Transformers versus convolutional architectures. \emph{Physical Review
E}, 112:065312.

Canatar, A., Bordelon, B., and Pehlevan, C. (2021). Spectral bias and
task-model alignment explain generalization in kernel regression and
infinitely wide neural networks. \emph{Nature Communications}, 12:2914.

Caponnetto, A. and De Vito, E. (2007). Optimal rates for the regularized
least-squares algorithm. \emph{Foundations of Computational
Mathematics}, 7(3):331--368.

Cheng, D., Liu, Z., Sun, J., Xia, F., Zhang, B., Liu, D., and Zhang, Y.
(2026). A qualitative test-risk mechanism for scaling behavior in
normalized residual networks. arXiv preprint, arXiv:2605.08297.

Chizat, L., Oyallon, E., and Bach, F. (2019). On lazy training in
differentiable programming. In \emph{Advances in Neural Information
Processing Systems 32 (NeurIPS)}, pages 2937--2947.

Defilippis, L., Krzakala, F., Loureiro, B., and Maillard, A. (2026a).
Optimal scaling laws in learning hierarchical multi-index models. arXiv
preprint, arXiv:2602.05846.

Defilippis, L., Xu, Y., Girardin, J., Troiani, E., Erba, V., Zdeborová,
L., Loureiro, B., and Krzakala, F. (2026b). Scaling laws and spectra of
shallow neural networks in the feature learning regime. In \emph{The
Fourteenth International Conference on Learning Representations (ICLR)}.

Hestness, J., Narang, S., Ardalani, N., Diamos, G., Jun, H., Kianinejad,
H., Patwary, M. M. A., Yang, Y., and Zhou, Y. (2017). Deep learning
scaling is predictable, empirically. arXiv preprint, arXiv:1712.00409.

Hoffmann, J., Borgeaud, S., Mensch, A., Buchatskaya, E., Cai, T.,
Rutherford, E., de Las Casas, D., Hendricks, L. A., Welbl, J., Clark,
A., Hennigan, T., Noland, E., Millican, K., van den Driessche, G.,
Damoc, B., Guy, A., Osindero, S., Simonyan, K., Elsen, E., Vinyals, O.,
Rae, J. W., and Sifre, L. (2022). Training compute-optimal large
language models. In \emph{Advances in Neural Information Processing
Systems 35 (NeurIPS)}, pages 30016--30030.

Huang, J., Wurgaft, D., Bansal, R., Ruis, L., Saphra, N., Alvarez-Melis,
D., Lampinen, A. K., Potts, C., and Lubana, E. S. (2026). Why larger
models learn more: Effects of capacity, interference, and rare-task
retention. arXiv preprint, arXiv:2605.29548.

Jha, N. K. and Reagen, B. (2026). Same architecture, different capacity:
Optimizer-induced spectral scaling laws. arXiv preprint,
arXiv:2605.21803.

Jelassi, S., Brandfonbrener, D., Kakade, S. M., and Malach, E. (2024).
Repeat after me: Transformers are better than state space models at
copying. In \emph{Proceedings of the 41st International Conference on
Machine Learning (ICML)}, PMLR 235:21502--21521.

Kaplan, J., McCandlish, S., Henighan, T., Brown, T. B., Chess, B.,
Child, R., Gray, S., Radford, A., Wu, J., and Amodei, D. (2020). Scaling
laws for neural language models. arXiv preprint, arXiv:2001.08361.

Li, J., Gu, X., Dai, R., Hao, X., Xu, C., Wu, Y., Zheng, S., and Zhang,
J. (2026). Scaling domain data repetition in LLM pretraining. arXiv
preprint, arXiv:2608.14071.

Liu, A. Z., Paquette, E., and Sous, J. (2026). Spectral lens: Activation
and gradient spectra as diagnostics of LLM optimization. arXiv preprint,
arXiv:2605.05683.

Liu, E., Sun, K., Li, M., Lee, I., Tjuatja, L., Huang, J.-T., and
Neubig, G. (2026). What do language models learn and when? The implicit
curriculum hypothesis. Accepted at the Conference on Language Modeling
(COLM 2026). arXiv:2604.08510.

Liu, Y. and Gore, J. (2026). Neural scaling universality: If exponents
are fixed, time to understand coefficients. arXiv preprint,
arXiv:2606.25008.

Liu, Y., Liu, Z., and Gore, J. (2025). Superposition yields robust
neural scaling. In \emph{Advances in Neural Information Processing
Systems 38 (NeurIPS)}. arXiv:2505.10465.

Maloney, A., Roberts, D. A., and Sully, J. (2022). A solvable model of
neural scaling laws. arXiv preprint, arXiv:2210.16859.

Michaud, E. J., Liu, Z., Girit, U., and Tegmark, M. (2023). The
quantization model of neural scaling. In \emph{Advances in Neural
Information Processing Systems 36 (NeurIPS)}, pages 28699--28722.

Ngo, K. and Ravanbakhsh, S. (2026). Scaling laws and symmetry, evidence
from neural force fields. In \emph{The Fourteenth International
Conference on Learning Representations (ICLR)}.

Nikolaou, K., Scheunemann, J., Krippendorf, S., Tovey, S., and Holm, C.
(2026). Spectral reach: Understanding neural scaling as progress into
the spectral tail. arXiv preprint, arXiv:2605.31244.

Ramani, V. and Jain, S. V. (2026). On the optimizer dependence of neural
scaling laws. In \emph{4th Workshop on High-dimensional Learning
Dynamics (HiLD), ICML 2026}. arXiv:2605.29387.

Schaeffer, R., Miranda, B., and Koyejo, S. (2023). Are emergent
abilities of large language models a mirage? In \emph{Advances in Neural
Information Processing Systems 36 (NeurIPS)}.

Sharma, U. and Kaplan, J. (2022). Scaling laws from the data manifold
dimension. \emph{Journal of Machine Learning Research}, 23(9):1--34.

Song, Z., Ji, S., Li, H., Cheng, S., and Huang, C. (2026). Data scaling
as progressive coverage of a predictive contribution spectrum. arXiv
preprint, arXiv:2605.20196.

Tay, Y., Dehghani, M., Abnar, S., Chung, H. W., Fedus, W., Rao, J.,
Narang, S., Tran, V. Q., Yogatama, D., and Metzler, D. (2023). Scaling
laws vs model architectures: How does inductive bias influence scaling?
In \emph{Findings of the Association for Computational Linguistics:
EMNLP 2023}, pages 12342--12364.

Volkova, A., Safaryan, M., Lampert, C. H., and Alistarh, D. (2026).
Towards robust scaling laws for optimizers. arXiv preprint,
arXiv:2602.07712.

Wang, S., Zhang, G., Luo, K., Wu, Y., Liu, S., Liu, J., Huang, W., Yan,
S., and Li, J. (2026). SMELT: Scaling laws for compute-matched MoE
looped Transformers. arXiv preprint, arXiv:2609.01343.

Xiao, C., Cai, J., Zhao, W., Lin, B., Zeng, G., Zhou, J., Zheng, Z.,
Han, X., Liu, Z., and Sun, M. (2025). Densing law of LLMs. \emph{Nature
Machine Intelligence}, 7:1823--1833.

Yang, G. and Hu, E. J. (2021). Tensor programs IV: Feature learning in
infinite-width neural networks. In \emph{Proceedings of the 38th
International Conference on Machine Learning (ICML)}, PMLR
139:11727--11737.

Zhang, J., Liu, Z., Yan, Z., Zhang, Y., Tan, G., Liu, F., and Cheng, D.
(2026). Mechanisms of width scaling in normalized residual networks: The
effective alignment dimension. arXiv preprint, arXiv:2607.24887.

Zou, J., Gong, Z., Su, Y., Tang, H., and Liu, Y. (2026). Effective
frontiers: A unification of neural scaling laws. arXiv preprint,
arXiv:2602.02593.

\end{document}